\pdfoutput=1
\documentclass[sigconf,nonacm]{acmart}
\AtBeginDocument{%
  }

\setcopyright{acmlicensed}
\copyrightyear{2027}
\acmYear{2027}
\acmDOI{XXXXXXX.XXXXXXX}
\acmConference[WWW '27]{The ACM Web Conference 2027}{May 10 to 14, 2027}{Dublin, Ireland}
\acmISBN{978-1-4503-XXXX-X/2018/06}

\usepackage{subcaption}
\usepackage{graphicx}
\usepackage[figuresright]{rotating}
\usepackage{multirow} %
\usepackage{tabularx} %
\usepackage{color}
\usepackage{colortbl}
\usepackage{enumitem} 
\usepackage[linesnumbered,ruled,vlined]{algorithm2e}
\newtheorem{tmremark}{Remark}

\begin{document}

\title[The Future Is Finer: Multivariate Time Series Super-Resolution Forecasting]{The Future Is Finer: Multivariate Time Series \\ Super-Resolution Forecasting}

\author{Shuhao Li}
\affiliation{%
  \institution{ Fudan University}
  \city{Shanghai}
  \country{China}}
  \affiliation{%
  \institution{Nanyang Technological University}
   \country{Singapore}}
\email{shli23@m.fudan.edu.cn}

\author{Weidong Yang}
\affiliation{%
  \institution{Fudan University}
  \city{Shanghai}
  \country{China}}
\affiliation{%
  \institution{Zhuhai Fudan Innovation Research Institute }
  \city{Zhuhai}
  \country{China}}

\author{Yue Cui}
\affiliation{%
  \institution{Tongyi Laboratory \\ Alibaba Group}
  \city{Hangzhou}
  \country{China}}

\author{Fan Zhang}
\affiliation{%
  \institution{Guangzhou University}
  \city{Guangzhou}
  \country{China}}

\author{Lu Qin}
\affiliation{%
  \institution{ University of Technology Sydney}
  \city{Sydney}
  \country{Australia}}

\renewcommand{\shortauthors}{Li et al.}

\begin{abstract}
Web platforms run on multivariate time series, from service telemetry and Web traffic logs to sensor streams served through Web APIs. Yet, retention policies often archive them at coarse granularity to save storage and bandwidth. Decisions such as autoscaling, alerting, and billing operate at finer granularity than existing forecasters reach, since these models predict at the resolution of their input. We therefore introduce \textbf{M}ultivariate \textbf{T}ime \textbf{S}eries \textbf{S}uper-\textbf{R}esolution \textbf{F}orecasting (\textbf{MTS-SRF}), which forecasts fine-grained futures from coarse histories, and propose \textbf{TimeMagnifier} to address its three challenges. Fourier Anchor counters data sparsity by fitting each input window to a trend-Fourier structure with sparse cross-variable gating; De-noise Only constrains fine-grained ambiguity by refining along the resolution axis rather than the noise axis, lengthening the grid level by level; and banded DTW Alignment corrects cross-resolution misalignment against the coarse forecast. On the first MTS-SRF benchmark, with ten datasets from five domains, real archived cross-resolution pairs, strengthened baseline adaptations, and five-seed significance tests, TimeMagnifier is best on every dataset, resolution coefficient, and metric against 15 baselines and cuts MSE by 12.3\% on average. Its De-noise Only and DTW stages also transfer as a plug-in that lowers the MSE of existing forecasters by 5.1\% to 9.0\%. Code will be made publicly available.
\end{abstract}


\maketitle

\section{Introduction}\label{sec:intro}

\begin{figure}[t]
\vspace{0.3cm}
\centering
\includegraphics[width=0.95\linewidth]{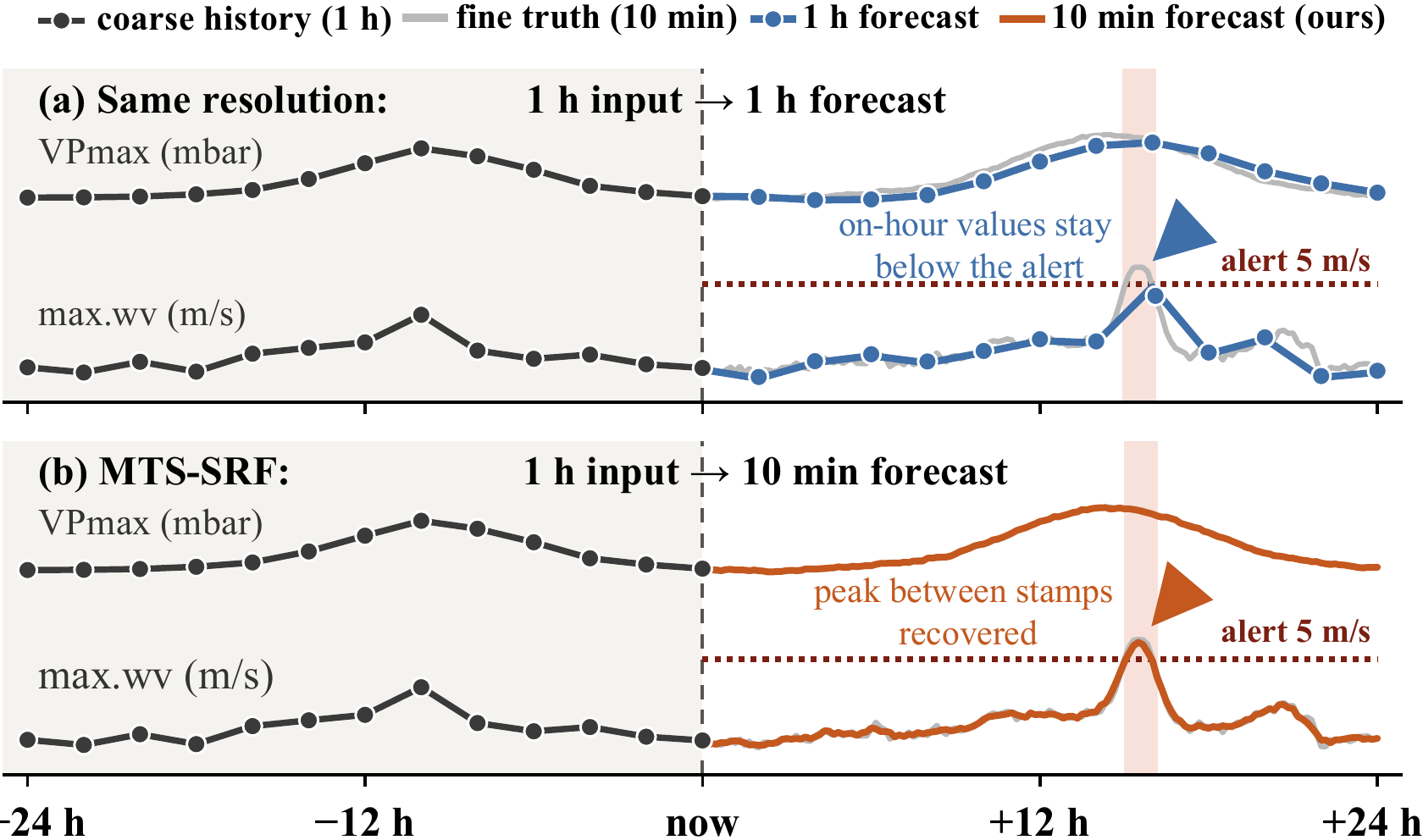}
        \vspace{-0.3cm}
\caption{Same-resolution forecasting versus MTS-SRF from the same hourly history. The hourly forecast returns only on-hour values, while the 10-minute forecast recovers the wind crest between them that crosses the alert level.}
\label{fig1}
\vspace{-0.5cm}
\end{figure}
Multivariate time series underpin many Web services, from server and network telemetry and Web traffic logs to environmental and energy sensors published through Web platforms \cite{runge2019inferring, avila2020data, march2021tracking}. They are collected at fixed intervals such as every minute, hour, or day \cite{zhou2021informer, lai2018modeling}, and to bound storage and transmission cost they are commonly rolled up to coarser intervals as they age, as in the downsampling retention of Web monitoring stacks \cite{adams2020monarch}. Existing forecasters, however, predict at the resolution of their input \cite{oreshkinn, cirstea2022triformer}, so a coarse archive cannot directly support decisions made at finer granularity.

For instance, when a Web-based weather service trains its forecasting model on hourly sampled data, the model cannot capture short-term variations, such as when rainfall will stop within the hour, which often matter more for real-world decisions. \textbf{Figure~\ref{fig1}} illustrates this on a Weather test window. From the same hourly history of saturated vapor pressure (VPmax) and maximum wind speed (max.wv), an hourly forecast returns only on-hour values that stay below the 5 m/s wind alert, whereas a 10-minute forecast recovers the crest between two stamps that crosses it, a slow swing below the coarse Nyquist frequency. The same gap arises on other Web platforms, where energy platforms schedule load in 15-minute intervals while archiving hourly records, and navigation services need hourly congestion forecasts from coarser traffic archives. Simply raising the sampling frequency is costly: (1) \textbf{higher collection cost}, as sensors and agents must report more often; (2) \textbf{storage and transmission that grow linearly with the sampling rate} across every monitored series; and (3) \textbf{greater computational overhead} for retraining~\cite{whang2020data,wang2020towards,ekambaram2023tsmixer,lin2024sparsetsf}.

Instead, we formalize multivariate time series super-resolution forecasting (MTS-SRF), which forecasts a finer future grid from coarse history alone and uses fine data only as training supervision, a combination that mixed-frequency, multi-scale, and super-resolution methods each relax. MTS-SRF is also not forecasting followed by interpolation, since interpolating a coarse forecast cannot by itself restore the between-stamp variation shown in \textbf{Figure~\ref{fig1}}. It poses three challenges. First, \textbf{data sparsity} limits the model’s ability to capture critical trends and patterns \cite{vandal2017deepsd,kuleshov2017audio}. Coarse data often lack essential details, and their noise can be magnified in fine-grained forecasts, leading to biased trend estimations. Second, \textbf{fine-grained ambiguity} complicates prediction, since many fine-grained futures are compatible with the same coarse history. Coarse-grained patterns reflect broad time scales and fail to capture short-term dynamics \cite{ghysels2007midas,koutsoyiannis2007statistical}, which matter most in time-critical tasks like real-time weather or traffic forecasting. Finally, \textbf{cross-resolution misalignment} disrupts modeling. A forecast refined on a finer grid can drift in phase against the coarse evidence, and real archives add timestamp jitter. Effective forecasting requires alignment across resolutions to prevent error propagation.

To address these challenges, we propose \textbf{TimeMagnifier}, built on a single idea, treating the coarse forecast as a resolution-degraded version of the fine future and refining it along the resolution axis, much as diffusion models refine along the noise axis. The \textit{Fourier Anchor} module counters data sparsity by fitting each input window to a trend-Fourier structure with sparse cross-variable gating, yielding a structured coarse forecast and per-window periodic priors. The \textit{De-noise Only} module constrains fine-grained ambiguity by lengthening the grid level by level, smoothing expansion artifacts and re-injecting these priors under fine supervision resampled to every level. The \textit{DTW Alignment} module corrects cross-resolution misalignment by realigning each level to the coarse forecast, thereby keeping refinement anchored. A multi-granularity loss trains all three stages jointly.
Our main contributions are as follows:
\begin{itemize}[leftmargin=*, itemsep=2pt, topsep=2pt]
\item We formulate the MTS-SRF problem, where models forecast high-resolution sequences from long-interval observations, addressing limitations of conventional fixed-resolution forecasting.
\item We propose \textbf{TimeMagnifier}, whose De-noise Only stage refines along the resolution axis rather than the noise axis, lengthening the grid at each level under resampled fine supervision and per-window periodic priors, and is coupled with banded DTW that realigns every level to the coarse forecast.
\item We show that the De-noise Only and DTW stages transfer to other forecasters, consistently improving DLinear and PatchTST when appended to their unchanged backbones.
\item We build the first MTS-SRF benchmark for deep multivariate forecasting on ten datasets from five domains, including real archived cross-resolution pairs, and show significant gains over strong and strengthened baselines under five seeds.
\end{itemize}

\section{Problem Formulation}

\begin{figure}[t]
\vspace{-0.2cm}
	\centering
		\includegraphics[width=1\linewidth]{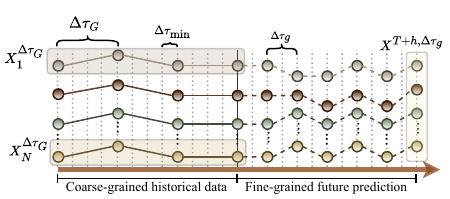}
        \vspace{-0.7cm}
		\caption{An example of the multivariate time series super-resolution forecasting problem, where the low-resolution coefficient is $ \Gamma = 4 $, i.e., $ \Delta \tau_G = 4 \Delta \tau_{\text{min}} $, and the high-resolution coefficient is $ \gamma = 2 $, i.e., $ \Delta \tau_{g} = 2 \Delta \tau_{\text{min}} $.}
		\label{fig_target pre}
        \vspace{-0.5cm}
\end{figure}

As illustrated in \textbf{Figure~\ref{fig_target pre}}, given a sampling time point \( t \), let \( x_i^{t, \Delta \tau_G} \) denote the observed value at sampling time \( t \) on the \( i \)-th time series with granularity \( \Delta \tau_G \), where \( \Delta \tau_G = \Gamma \Delta \tau_{\text{min}} \), \( \Delta \tau_{\text{min}} \) is the smallest sampling interval, i.e., the highest resolution, and \( \Gamma \) is the low-resolution coefficient, indicating the factor by which the sampling interval of the coarse-grained time series exceeds the highest resolution \( \Delta \tau_{\text{min}} \). Specifically, a time series with granularity \( \Delta \tau_G \) has a longer sampling period than \( \Delta \tau_{\text{min}} \), and \( \Gamma \) is a positive integer representing the magnification of the lower resolution relative to the highest resolution.

Thus, the set \( X^{t, \Delta \tau_G} = \{ x_1^{t, \Delta \tau_G}, x_2^{t, \Delta \tau_G}, \dots, x_N^{t, \Delta \tau_G} \} \) represents the observed values of \( N \) variables at granularity \( \Delta \tau_G \) and time point \( t \).

\noindent\textbf{Multivariate Time Series Super-Resolution Forecasting (MTS-SRF)}. Given a historical look-back time window \( T \), let \( X = \{ X^{1, \Delta \tau_G},\allowbreak\, X^{2, \Delta \tau_G}, \dots,\, X^{T, \Delta \tau_G} \} \) represent the observed values of the low resolution time series \( X \) over the past \( T \) time steps, where each \( X^{t, \Delta \tau_G} \) is the $N$-variate observation with granularity \( \Delta \tau_G \) at time \( t \). Given a requested target-grid horizon $h_{\mathrm{req}}$, let $\Gamma,\gamma\in\mathbb{N}^{+}$ with $\gamma\leq\Gamma$, $d = \gcd(\Gamma,\gamma)$, $p = \Gamma/d$, and $q = \gamma/d$. To keep both grids integral and temporally compatible for any rational ratio, we retain complete resolution blocks:
\begin{equation}
m = \left\lceil \frac{h_{\mathrm{req}}}{p} \right\rceil, \quad h = mp, \quad h_G = mq, \quad r = \frac{h}{h_G} = \frac{\Gamma}{\gamma} .
\label{eq:compatible_horizon}
\end{equation}
Here \( \Delta \tau_{g} = \gamma \Delta \tau_{\text{min}} \) with \( \gamma \) the high-resolution coefficient; the at most $p-1$ extra steps use real targets and are not scored. The primary MTS-SRF setting is \( \gamma < \Gamma \), while \( \gamma = \Gamma \) reduces to same-resolution forecasting.

\begin{figure*}[t]
\vspace{-0.3cm}
	\centering
		\includegraphics[width=1\linewidth]{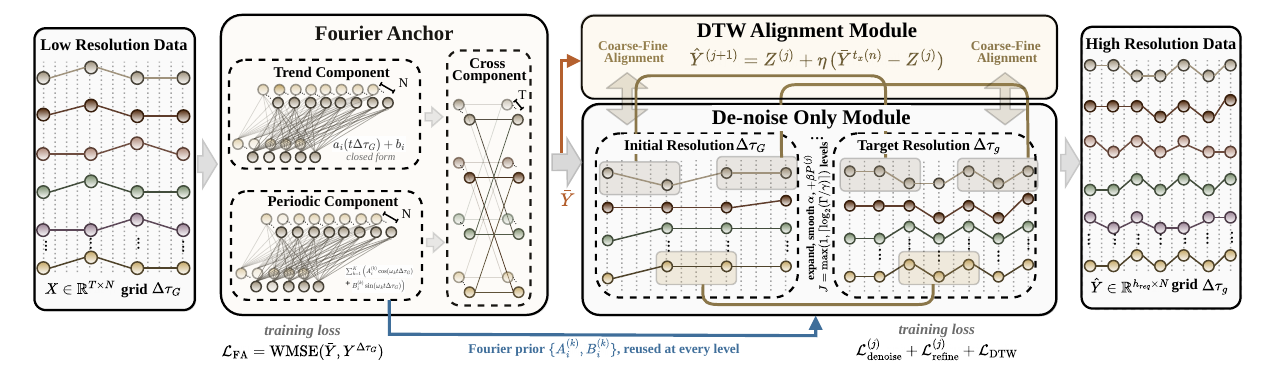}
  \vspace{-0.9cm}
		\caption{Illustration of TimeMagnifier's architectural framework and inference data flow with \( \Gamma = 4, \gamma = 2 \) as an example.}
		\label{fig: RoadDiff}
  \vspace{-0.3cm}
\end{figure*}

\noindent\textbf{Objective:} The goal is to generate future fine-grained data from historical low-resolution data, similar to super-resolution restoration in computer vision \cite{lei2024dvmsr,wang2020deep}.

\noindent\textbf{Remark on supervision.} MTS-SRF requires fine-resolution data only as training supervision; at deployment the model consumes coarse history alone, matching settings where fine-grained collection is limited by cost, storage, or sensors. The benchmark therefore constructs coarse inputs by point-sampling, as image super-resolution does with synthetic degradation. ETT also offers real archived pairs, since its official hourly ETTh files are exactly the on-hour records of the 15-minute ETTm files, and the $\Gamma=4$ ETTm columns of \textbf{Table~\ref{tab: main}} use these pairs.

\noindent\textbf{Formal Expression.} We seek a mapping $f$ from the low-resolution history $X$ to the future high-resolution series $\hat{Y}$, that is:
\begin{equation}
    \hat{Y} = f(X; \theta)
\end{equation}
where $f(\cdot; \theta)$ represents the forecasting function, $\theta$ are the model parameters, and $\hat{Y}$ collects the first $h_{\mathrm{req}}$ of the $h$ internal steps, estimating the ground-truth high-resolution series $Y = \{ Y^{1, \Delta\tau_g}, \dots,\allowbreak Y^{h_{\mathrm{req}}, \Delta\tau_g}\} \in \mathbb{R}^{h_{\mathrm{req}} \times N}$, with coarse counterpart $Y^{\Delta\tau_G}$.

\section{TimeMagnifier Framework}

To tackle MTS-SRF, we propose TimeMagnifier, a lightweight coarse-to-fine framework illustrated in \textbf{Figure~\ref{fig: RoadDiff}}. TimeMagnifier decomposes super-resolution forecasting into three coupled stages that mirror the three challenges identified in Section~\ref{sec:intro}. The \textit{Fourier Anchor} stage counteracts downsampling-induced information loss by estimating the trend and periodic structure that survive coarse sampling and borrowing complementary information from correlated variables, producing a structured coarse forecast \( \bar{Y} \) together with Fourier coefficients \( \{A_i^{(k)}, B_i^{(k)}\} \). The \textit{De-noise Only} stage constrains fine-grained ambiguity. Instead of predicting the fine sequence in one shot, it decomposes the \( \Gamma/\gamma \) resolution gap into successive refinement levels, reusing the Fourier coefficients of the first stage as periodic priors to constrain the ill-posed upsampling. The \textit{DTW Alignment} stage corrects cross-resolution misalignment by warping the refined sequence back onto the coarse prior \( \bar{Y} \), so that every refinement step remains temporally consistent with the structured coarse forecast. The three stages are output-coupled rather than concatenated, since the second reuses the periodic coefficients of the first, the third aligns against the output of the first, and a multi-granularity loss optimizes all of them jointly.

\subsection{Fourier Anchor Module}

Coarse sampling removes high-frequency detail, but the components that survive it, namely long-term trends, dominant periodicities, and cross-variable correlations, are exactly what a super-resolution forecaster can still rely on \cite{shao2022spatial,zeng2023transformers,deng2021st}. The \textit{Fourier Anchor} stage therefore makes this surviving structure explicit by decomposing each variable into a linear trend, a Fourier-based periodic term, and a learnable cross-series interaction term, yielding both a stable coarse-grained forecast at granularity \( \Delta \tau_G \) and periodic coefficients that the subsequent stage reuses at finer resolutions.

\noindent\textbf{Input-Conditioned Trend and Periodic Components.}
Coarse sampling preserves the linear trend and the dominant periodicities below the coarse Nyquist frequency, which Fourier series theory lets us represent by sines and cosines \cite{jenkins2022fourier}. We collect both into the basis
\begin{equation}
\phi(\tau) = \left[ 1, \, \tau, \, \cos(\omega_1 \tau), \, \sin(\omega_1 \tau), \dots, \cos(\omega_K \tau), \, \sin(\omega_K \tau) \right]^\top
\end{equation}
with $\omega_k = 2\pi\Gamma\Delta\tau_{\text{min}} / \Pi_k$ for $K$ dominant periods $\Pi_k$ fixed on the coarse training split, and write $\Phi_H \in \mathbb{R}^{T \times D}$ and $\Phi_F \in \mathbb{R}^{h_G \times D}$, $D = 2 + 2K$, for the basis matrices evaluated on the historical grid $\tau_t^{H} = t-T$ and the future coarse grid $\tau_n^{F} = n$, both expressed in units of $\Delta\tau_G$, so that the last observed step sits at time zero and the basis is dimensionless. For each input $X$, Fourier Anchor fits the coefficients by ridge regression and extrapolates them forward:
\begin{equation}
C(X) = \left( \Phi_H^\top \Phi_H + \lambda_{\mathrm{ridge}} R \right)^{-1} \Phi_H^\top X , \qquad \widetilde{Y}(X) = \Phi_F \, C(X) ,
\label{eq:conditional_Fourier Anchor}
\end{equation}
where $R = \operatorname{diag}(0,1,\dots,1)$ leaves the intercept unpenalized. For fixed frequencies this is an explicit linear map $\widetilde{Y}(X) = P_{\mathrm{FA}} X$ with $P_{\mathrm{FA}} = \Phi_F ( \Phi_H^\top \Phi_H + \lambda_{\mathrm{ridge}} R )^{-1} \Phi_H^\top$, shared across variables and independent of $N$. The column $c_i(X)$ of $C(X)$ holds the intercept, trend, and Fourier coefficients $\{b_i(X), a_i(X), A_i^{(k)}(X), B_i^{(k)}(X)\}$ of variable $i$, ordered as in $\phi(\tau)$, which are reused as input-conditioned periodic priors during refinement.

\noindent\textbf{Cross Component.}
Channel-dependent strategies can outperform channel-independent ones but incur higher computational cost \cite{qiu2025comprehensive,liu2024itransformer}. We therefore adopt a sparse shared-gating design that retains effectiveness on a compact parameter budget:
\begin{equation}
a_{ij}^{n} = \frac{\exp(g_{ij}^{n})}{\sum_{\ell \in \mathcal{N}_i} \exp(g_{i\ell}^{n})}, \qquad \bar{Y}_i^{n}(X) = \widetilde{Y}_i^{n}(X) + \kappa \sum_{j \in \mathcal{N}_i} a_{ij}^{n} \cdot \widetilde{Y}_j^{n}(X)
\end{equation}
where $g_{ij}^n = \text{MLP}_{\theta_g}([\widetilde{Y}_i^{n}(X); \widetilde{Y}_j^{n}(X)])$ is a shared lightweight network learning the interaction strength between variables $i$ and $j$ at coarse step $n$, $\mathcal{N}_i$ collects the $|\mathcal{N}_i| = \min(16, N-1)$ channels most correlated with $i$ under absolute Pearson correlation on the coarse training split, and $\kappa$ is a learnable scalar controlling the strength of the cross-variable residual. Since both $\widetilde{Y}$ and the gates are functions of $X$, the coarse forecast $\bar{Y}$ is fully input-conditioned. We abbreviate $\bar{Y}_i^{n}(X)$ as $\bar{Y}_i^{n}$ in what follows.

\noindent\textbf{Optimization.}\label{subsec:cl_opt}
The trend and Fourier coefficients are computed from each input through Eq.~\eqref{eq:conditional_Fourier Anchor} and are not global model parameters; the trainable Fourier Anchor parameters are the shared interaction MLP $\theta_g$ and the residual scalar $\kappa$, while $\lambda_{\mathrm{ridge}} = 10^{-3}$ is fixed across all datasets and resolution settings. They are trained by minimizing a weighted mean squared error (WMSE):
\begin{equation}
    \mathcal{L}_{\text{FA}} = \sum_{i=1}^{N} \varpi_i \left( \frac{1}{h_G} \sum_{n=1}^{h_G} \left( \bar{Y}_i^{n}(X) - Y_i^{n, \Delta \tau_G} \right)^2 \right)
\end{equation}
where the series weights $\varpi_i$ default to uniform and $Y_i^{n, \Delta \tau_G}$ is the ground-truth observation at granularity $\Delta \tau_G$. Because $\widetilde{Y}(X) = \Phi_F C(X)$, the coarse forecast lies in the trend-Fourier subspace $\operatorname{col}(\Phi_F)$ before cross-variable mixing. At both training and inference, $\bar{Y}$ and $\{ A_i^{(k)}(X), B_i^{(k)}(X) \}$ are passed to the De-noise Only Module as its initialization and periodic prior.

\subsection{De-noise Only Module}
Even with the surviving structure recovered, upsampling a coarse forecast onto a finer grid remains ill-posed, since many fine-grained futures are compatible with the same coarse estimate. Inspired by the coarse-to-fine refinement principle of diffusion-based generative models~\cite{rasul2021autoregressive, ho2020denoising}, we treat the Fourier Anchor output as a resolution-degraded version of the fine-grained target and refine it deterministically, bypassing stochastic noise injection. Starting from \(\hat{Y}^{(0)} = \bar{Y}\), the \textit{De-noise Only} stage applies \( J \) refinement steps, each of which refines the sampling interval, smooths the expanded sequence, and re-injects the periodic prior learned by Fourier Anchor, transforming \(\hat{Y}^{(0)}\) into a fine-grained sequence \(\hat{Y}^{(J)}\).

\noindent\textbf{Granularity Expansion.}
For the boundary case $r=1$ we set $J=1$ with identity expansion, so the output still passes one level of smoothing, periodic correction, and alignment. The refinement ladder is defined on sequence lengths: let $L^{(0)} = h_G$ and
\begin{equation}
L^{(j+1)} = \min \left( 2 L^{(j)}, \, h \right), \qquad J = \max\big(1, \left\lceil \log_2 r \right\rceil\big) ,
\label{eq:length_schedule}
\end{equation}
so the discrete expansion factor $\tilde{q}_j = L^{(j+1)} / L^{(j)} \in [1,2]$ equals two except possibly at the final level, and one only when $r=1$. Because $h = r h_G$ by Eq.~\eqref{eq:compatible_horizon}, the normalized sampling interval at level $j$ is
\begin{equation}
\Delta \tau^{(j)} = \frac{h_G}{L^{(j)}} ,
\end{equation}
so that $\Delta \tau^{(0)} = 1$ and $\Delta \tau^{(J)} = 1/r$ hold exactly in coarse-step units. We index the level $j$ grid $\mathcal{G}^{(j)}$ by $n = 1, \dots, L^{(j)}$, with interval-end timestamps $\tau_n^{(j)} = n \Delta \tau^{(j)}$, matching the future coarse grid $\tau_n^{F}$; expansion is a resampling operator onto the next grid:
\begin{equation}
    \hat{Y}_i^{(j,1)} = \mathcal{I}_{\mathcal{G}^{(j)} \to \mathcal{G}^{(j+1)}} \big( \hat{Y}_i^{(j)} \big)
\end{equation}
which reduces to value replication when $\tilde{q}_j = 2$ and to nearest-neighbor resampling when $1 < \tilde{q}_j < 2$. The grid lengths double from $h_G$, with a possibly shorter final expansion that reaches $h$, so the total refined length is $O(h)$.

\noindent\textbf{Denoising via Local Smoothing.}\label{subsec:smoothing}
Following granularity expansion, the interpolated sequence may introduce artificial noise or abrupt transitions. To address this, we apply a smoothing operation that preserves local continuity and global trend consistency:
\begin{equation}
\begin{aligned}
    \hat{Y}_i^{(j,2)}(n) & = \hat{Y}_i^{(j,1)}(n) - \alpha \Big(  \hat{Y}_i^{(j,1)}(n) \\
    & - \tfrac{1}{2} \big( \hat{Y}_i^{(j,1)}(n-1) + \hat{Y}_i^{(j,1)}(n+1) \big) \Big)
\end{aligned}
\end{equation}
where the boundary neighbors are defined by replication,
\begin{equation}
\hat{Y}_i^{(j,1)}(0) = \hat{Y}_i^{(j,1)}(1), \qquad \hat{Y}_i^{(j,1)}\big(L^{(j+1)}+1\big) = \hat{Y}_i^{(j,1)}\big(L^{(j+1)}\big),
\label{eq:smoothing_boundary}
\end{equation}
which avoids artificial wrap-around and writes the update as $S_\alpha = I - \alpha \mathbf{L}$ with $\mathbf{L}$ symmetric positive semidefinite. The learnable scalar $\alpha = \sigma(a) \in (0,1)$ is shared across levels, with $a$ initialized to zero, so $\alpha$ stays in the range required by Remark~\ref{prop:converge}.

\noindent\textbf{Periodic Correction.}
To restore the periodic structure learned by Fourier Anchor at the finer resolution, we reuse the Fourier coefficients \( \{ A_i^{(k)}, B_i^{(k)} \} \) and define:
\begin{equation}
\begin{aligned}
p_i(\tau) &= \sum_{k=1}^{K} \left( A_i^{(k)}(X) \cos(\omega_k \tau) + B_i^{(k)}(X) \sin(\omega_k \tau) \right), \\
P_i^{(j)}(n) &= p_i\big(\tau_n^{(j+1)}\big) - p_i\big(\tau_{\rho_j(n)}^{(j)}\big),
\end{aligned}
\end{equation}
where $\rho_j(n)$ is the level-$j$ index that expansion copies to $n$, and $A_i^{(k)}(X)$ and $B_i^{(k)}(X)$ are the periodic amplitudes at frequency $\omega_k$ estimated from the current window by Eq.~\eqref{eq:conditional_Fourier Anchor}, so the prior injected at each level is input-conditioned rather than a globally shared template. Since expansion already replicates the coarse-grid periodic values, we add only the attenuated in-between residual:
\begin{equation}
\hat{Y}_i^{(j,3)}(n) = \hat{Y}_i^{(j,2)}(n) + \beta P_i^{(j)}(n)
\label{eq:periodic_corr}
\end{equation}
where $\beta = \sigma(b) \in (0,1)$ is a learnable scalar periodic attenuation coefficient shared across all refinement levels, with $b$ initialized to zero. This step injects periodic variations as a controlled prior, preventing over-amplification while maintaining consistency with the coarse-grained temporal context.

\noindent\textbf{Iterative Refinement.}
Each level applies expansion, one smoothing step, one periodic correction, and the DTW correction of \textbf{Section~\ref{subsec:dtw_correct}}, mapping $\hat{Y}^{(j)}$ to $\hat{Y}^{(j+1)}$. All $J$ levels are executed without early stopping, so the output is always defined on the target grid $\Delta\tau_g$. A single smoothing pass attenuates the high-frequency artifacts introduced by expansion and, by Remark~\ref{prop:converge}, never amplifies the sequence. After all $J$ levels, the final sequence $\hat{Y}^{(J)}$ is obtained.

\subsection{DTW Alignment Module}
Refinement alone can drift, since on finer grids small per-step errors accumulate into phase shifts against the coarse evidence. The \textit{DTW Alignment} stage closes this loop by warping \( \hat{Y}^{(j)} \) back onto the structured coarse prior \( \bar{Y} \) produced by Fourier Anchor. We align to \( \bar{Y} \) rather than raw observations because future coarse values are unavailable at deployment; \( \bar{Y} \) is supervised by \( \mathcal{L}_{\text{FA}} \), and since, by Remark~\ref{prop:dtwbound}, the update interpolates between refinement and anchor errors, the step $\eta$, trained through the fine-grid loss, can learn a small value when the anchor is unreliable. It integrates classical DTW \cite{berndt1994using} with learnable weights, following the global structure of \( \bar{Y} \) while staying flexible locally.

\noindent\textbf{Temporal Alignment via Dynamic Time Warping.}
Given the structured coarse-grained sequence $\bar{Y} = \{ \bar{Y}_1^{t, \Delta \tau_G}, \dots, \bar{Y}_N^{t, \Delta \tau_G} \}$ and the pre-alignment sequence $Z^{(j)} = \hat{Y}^{(j,3)}$ on $\mathcal{G}^{(j+1)}$, the alignment objective is to minimize the cumulative distance between them:
\begin{equation}
D(\bar{Y}, Z^{(j)}) = \sum_{i=1}^{N} \min_{\pi_i^{(j)}} \sum_{(t_x, n) \in \pi_i^{(j)}} d\left( \bar{Y}_i^{t_x}, Z_i^{(j)}(n) \right)
\label{eq:dtw_cost}
\end{equation}
where $\pi_i^{(j)}$ is the alignment path for variable $i$ at level $j$ and $d(\cdot,\cdot)$ is the squared Euclidean distance on standardized values. To restrict the search around the scaled diagonal, each coarse index $t_x \in \{1,\dots,h_G\}$ is matched only against
\begin{equation}
\mathcal{B}_j(t_x) = \left\{ n : \left| n - \left( 1 + \frac{(t_x-1)(L^{(j+1)}-1)}{\max(h_G-1,1)} \right) \right| \le B \right\} ,
\label{eq:dtw_band}
\end{equation}
with half-width $B = \lceil r \rceil$, which also contains the interval-end match $n = \lceil t_x L^{(j+1)}/h_G \rceil$, so each level costs $O(Nh)$ instead of the $O(N h_G h)$ of unconstrained DTW.

\noindent\textbf{Learnable Alignment Weights for Adaptive Adjustment.} While traditional DTW weighs all cell pairs equally, it cannot emphasize the pairs that matter during refinement. To overcome this limitation, we introduce a learnable weighting function that reweights the matched pairs in the alignment loss. Given an input pair $(\bar{Y}_i^{t_x},\allowbreak Z_i^{(j)}(n))$, the alignment weight is computed as:
\begin{equation}
\begin{aligned}
    \tilde{\nu}_i^{(j)}(t_x, n) &= \operatorname{softplus}\big( \mathcal{A}_{\psi}( \bar{Y}_i^{t_x}, Z_i^{(j)}(n) ) \big), \\
    \nu_i^{(j)}(t_x, n) &= \tilde{\nu}_i^{(j)}(t_x, n) \Big/ \operatorname{mean}_{(u,v) \in \pi_i^{(j)}} \tilde{\nu}_i^{(j)}(u,v).
\end{aligned}
\end{equation}
where $\mathcal{A}_{\psi}$ is a learnable alignment network parameterized by $\psi$. The softplus keeps the weights positive and the mean over the cells of the path fixes their overall scale per variable, so the network expresses relative importance without being able to lower the loss by uniformly shrinking all weights. These weights define the weighted alignment cost used in the loss:
\begin{equation}
D'(\bar{Y}, Z^{(j)}) = \sum_{i=1}^{N} \sum_{(t_x, n) \in \pi_i^{(j)}} \nu_i^{(j)}(t_x, n)\, d\left( \bar{Y}_i^{t_x}, Z_i^{(j)}(n) \right)
\end{equation}
To rule out a cheap low-weight path, $\pi_i^{(j)}$ is chosen under the unweighted cost of Eq.~\eqref{eq:dtw_cost} and $\nu$ enters only the loss. Since the path search \( \min_{\pi_i^{(j)}} \) is non-differentiable, we fix the DP-computed path per forward pass with a stop-gradient and back-propagate only through \( \nu \), the sequence values, and \( \eta \), avoiding a soft relaxation.

\noindent\textbf{Refinement-Based Alignment Correction.}\label{subsec:dtw_correct}
At the end of each level, DTW warps the corrected sequence \( \hat{Y}^{(j,3)} \) toward \( \bar{Y} \):
\begin{equation}
\hat{Y}_i^{(j+1)}(n) = \hat{Y}_i^{(j,3)}(n) + \eta \left( \bar{Y}_i^{t_x(n)} - \hat{Y}_i^{(j,3)}(n) \right)
\end{equation}
where $t_x(n)$ is, among the coarse indices matched to $n$ on the DTW path, the one nearest in time (the larger on ties), initialized as $t_x(n) = \min\!\big(h_G,\, 1 + \lfloor (n-1) L^{(0)} / L^{(j+1)} \rfloor\big)$ and recomputed each level, and $\eta=\sigma(\xi)\in(0,1)$ with a single trainable $\xi$.

\noindent\textbf{Alignment Loss Function.}\label{subsec:dtw_loss}
While the alignment correction operates at every refinement step, the alignment loss is imposed on the final output \( \hat{Y}^{(J)} \), with the band of Eq.~\eqref{eq:dtw_band} taken at length $h$:
\begin{equation}
\mathcal{L}_{\text{DTW}} = \sum_{i=1}^{N} \sum_{(t_x, n) \in \pi_i^{(J)}} \nu_i^{(J)}(t_x, n) \left( \bar{Y}_i^{t_x} - \hat{Y}_i^{(J)}(n) \right)^2
\end{equation}
Minimizing \( \mathcal{L}_{\text{DTW}} \) encourages the refined outputs to stay close to the structured prior while adapting across granularities through learned alignment weights.

\subsection{Training Strategy}\label{subsec:training}

The three stages are optimized jointly rather than sequentially, and a multi-granularity loss supervises the coarse estimate, the intermediate refinements, and the cross-resolution alignment on the same forward pass, so that errors made at one resolution are corrected before they propagate to the next.

\noindent\textbf{Denoising and Refinement Loss.}
The De-noise Only Module iteratively smooths expansion artifacts and refines the time series. Its training process involves two loss functions:

\noindent\textit{Denoising Loss:} To enforce local smoothness, we define:
\begin{equation}
  \mathcal{L}_{\text{denoise}}^{(j)} = \sum_{i=1}^{N} \sum_{n=2}^{L^{(j)}-1} \left( \hat{Y}_i^{(j)}(n) - \frac{1}{2} \left( \hat{Y}_i^{(j)}(n-1) + \hat{Y}_i^{(j)}(n+1) \right) \right)^2
\end{equation}
which penalizes deviations from the local two-point average on the refined grid of step \( j \), encouraging smoother transitions.

\noindent\textit{Refinement Loss:} To ensure the predicted sequence converges toward the fine-grained target, we minimize:
\begin{equation}
  \mathcal{L}_{\text{refine}}^{(j)} = \sum_{i=1}^{N} \sum_{n=1}^{L^{(j)}} \left( \hat{Y}_i^{(j)}(n) - Y_i^{(j)}(n) \right)^2
\end{equation}
where $Y^{(j)}$ is the fine-resolution target resampled onto the level $j$ grid $\mathcal{G}^{(j)}$, so that supervision is applied at the resolution actually produced by that level; at $j=J$ it extends $Y^{\Delta\tau_g}$.

\noindent\textbf{Overall Optimization Objective.}
The Fourier Anchor objective $\mathcal{L}_{\text{FA}}$ of \textbf{Section~\ref{subsec:cl_opt}} enters unchanged. Writing $\mathcal{L}_2 = \sum_{j=1}^{J} ( \mathcal{L}_{\text{denoise}}^{(j)} + \mathcal{L}_{\text{refine}}^{(j)} )$ for the aggregated refinement objective and $\mathcal{L}_3 = \mathcal{L}_{\text{DTW}}$ for the alignment objective of \textbf{Section~\ref{subsec:dtw_loss}}, the total loss is
\begin{equation}
\mathcal{L}_{\text{total}} = e^{-s_1} \mathcal{L}_{\text{FA}} + s_1 + \sum_{m=2}^{3} \left( e^{-s_m} \mathcal{L}_m + s_m \right)
\label{eq:total}
\end{equation}
where the effective weights $\lambda_m = e^{-s_m} > 0$ balance trend supervision, refinement, and alignment. The additive $s_m$ penalize shrinking any weight, so the trivial all-zero solution is excluded, and the $s_m$ are learned jointly with $\theta$. The refinement and alignment objectives stay active when $r=1$, so both modules remain trained in the same-resolution case. All objectives are optimized in the same forward and backward pass, and their costs are analyzed in \textbf{Appendix~\ref{app:complexity}}.

\section{Experiments}
We conduct extensive experiments on multivariate time series benchmarks to evaluate TimeMagnifier and establish performance baselines for the MTS-SRF task. Full $\Gamma$ and $\gamma$ results and case visualizations will be released with the code.

\begin{table*}[!t]
\setlength\tabcolsep{2.5pt} 
  \centering
   
  \caption{Main results. The \textcolor[rgb]{ .871,  .737,  .604}{brown background} marks the best result and the \textcolor[rgb]{.349,.349,.349}{gray background} the second best.}
  \vspace{-0.3cm}
  \resizebox{0.98\linewidth}{!}{
    \begin{tabular}{ccccc|cc|cc|cc|cc|cc|cc|cc|cc|cc}
    \toprule
    \multirow{2}[4]{*}{\textbf{Category}} & \multirow{2}[4]{*}{\textbf{Models}} & \multirow{2}[4]{*}{\textbf{Metrics}} & \multicolumn{2}{c}{\textbf{BJAir}} & \multicolumn{2}{c}{\textbf{ETTh1}} & \multicolumn{2}{c}{\textbf{ETTh2}} & \multicolumn{2}{c}{\textbf{ETTm1}} & \multicolumn{2}{c}{\textbf{ETTm2}} & \multicolumn{2}{c}{\textbf{Electricity}} & \multicolumn{2}{c}{\textbf{ExRate}} & \multicolumn{2}{c}{\textbf{Illness}} & \multicolumn{2}{c}{\textbf{Traffic}} & \multicolumn{2}{c}{\textbf{Weather}} \\
\cmidrule{4-23}          &       &       & $\boldsymbol{\Gamma\!=\!2}$ & $\boldsymbol{\Gamma\!=\!4}$ & $\boldsymbol{\Gamma\!=\!2}$ & $\boldsymbol{\Gamma\!=\!4}$ & $\boldsymbol{\Gamma\!=\!2}$ & $\boldsymbol{\Gamma\!=\!4}$ & $\boldsymbol{\Gamma\!=\!2}$ & $\boldsymbol{\Gamma\!=\!4}$ & $\boldsymbol{\Gamma\!=\!2}$ & $\boldsymbol{\Gamma\!=\!4}$ & $\boldsymbol{\Gamma\!=\!2}$ & $\boldsymbol{\Gamma\!=\!4}$ & $\boldsymbol{\Gamma\!=\!2}$ & $\boldsymbol{\Gamma\!=\!4}$ & $\boldsymbol{\Gamma\!=\!2}$ & $\boldsymbol{\Gamma\!=\!4}$ & $\boldsymbol{\Gamma\!=\!2}$ & $\boldsymbol{\Gamma\!=\!4}$ & $\boldsymbol{\Gamma\!=\!2}$ & $\boldsymbol{\Gamma\!=\!4}$ \\
    \midrule
    \midrule
    \multirow{16}[16]{*}{\begin{sideways}\textbf{Transformer-based}\end{sideways}} & \multirow{2}[2]{*}{\textbf{Transformer}} & \textbf{MSE} & 0.9431 & 1.0685 & 1.1564 & 1.1454 & 1.7930 & 1.9874 & 1.6275 & 1.6526 & 1.1175 & 1.2153 & 0.4665 & 0.4923 & 1.2372 & 1.3889 & 4.3217 & 4.6351 & 1.1573 & 1.3446 & 1.1214 & 1.1643 \\
          &       & \textbf{MAE} & 0.7321 & 0.7888 & 0.9540 & 0.9808 & 1.1406 & 1.2586 & 1.2037 & 1.2204 & 0.9581 & 1.0350 & 0.6187 & 0.6604 & 0.9673 & 1.0488 & 1.7474 & 1.7802 & 0.6423 & 0.7376 & 0.8519 & 0.9214 \\
\cmidrule{3-23}          & \multirow{2}[2]{*}{\textbf{Informer}} & \textbf{MSE} & 1.0782 & 1.2093 & 1.1967 & 1.2143 & 1.9497 & 2.0843 & 1.6675 & 1.6983 & 1.2605 & 1.3879 & 0.5653 & 0.6526 & 1.3198 & 1.3937 & 4.5686 & 4.7428 & 1.5136 & 1.8341 & 1.1657 & 1.2052 \\
          &       & \textbf{MAE} & 0.7809 & 0.8269 & 0.9845 & 1.0378 & 1.1533 & 1.1882 & 1.1673 & 1.1773 & 0.9513 & 1.0995 & 0.6892 & 0.7548 & 0.8784 & 0.8955 & 1.6226 & 1.6495 & 0.8202 & 0.9478 & 0.8914 & 0.9503 \\
\cmidrule{3-23}          & \multirow{2}[2]{*}{\textbf{Autoformer}} & \textbf{MSE} & 0.8644 & 0.9352 & 0.6564 & 0.7307 & 0.8208 & 0.9106 & 0.8470 & 0.8552 & 0.5438 & 0.8276 & 0.4091 & 0.5110 & 0.9909 & 1.2998 & 3.6372 & 3.9121 & 1.0796 & 1.3081 & 0.5776 & 0.5855 \\
          &       & \textbf{MAE} & 0.7102 & 0.7400 & 0.6584 & 0.7405 & 0.8313 & 0.8807 & 0.8202 & 0.8267 & 0.6040 & 0.8635 & 0.5869 & 0.6760 & 0.8387 & 0.9857 & 1.5833 & 1.6930 & 0.6827 & 0.7889 & 0.6452 & 0.6530 \\
\cmidrule{3-23}          & \multirow{2}[2]{*}{\textbf{FEDformer}} & \textbf{MSE} & 0.9215 & 1.0064 & \cellcolor[rgb]{ .851,  .851,  .851} 0.6236 & \cellcolor[rgb]{ .851,  .851,  .851} 0.6811 & 0.8963 & 0.9315 & 0.6659 & 0.6797 & 0.5724 & 0.8793 & 0.3088 & 0.3243 & 1.0216 & 1.3097 & 3.2787 & 3.5748 & 0.8042 & 0.9414 & 0.4846 & 0.4920 \\
          &       & \textbf{MAE} & 0.7561 & 0.8311 & 0.6427 & 0.7041 & 0.8070 & 0.8582 & 0.7138 & 0.7291 & 0.5755 & 0.8818 & 0.5176 & 0.5195 & 0.7975 & 0.9487 & 1.5171 & 1.6526 & 0.5539 & 0.6450 & 0.5704 & 0.5791 \\
\cmidrule{3-23}          & \multirow{2}[2]{*}{\textbf{Crossformer}} & \textbf{MSE} & 0.7856 & 0.8390 & 0.7193 & 0.7877 & 1.3547 & 1.4588 & 0.7961 & 0.8039 & 1.0514 & 1.5604 & 0.3549 & 0.4452 & 1.2961 & 1.3994 & 3.6996 & 3.8804 & 0.7226 & 0.8756 & 0.3889 & 0.4069 \\
          &       & \textbf{MAE} & 0.6797 & 0.7001 & 0.7150 & 0.7837 & 0.9823 & 1.1083 & 0.8008 & 0.8071 & 0.8522 & 1.1866 & 0.5302 & 0.6177 & 0.9576 & 1.0365 & 1.5908 & 1.7399 & 0.4411 & 0.5097 & 0.5029 & 0.5561 \\
\cmidrule{3-23}          & \multirow{2}[2]{*}{\textbf{PatchTST}} & \textbf{MSE} & 0.7273 & 0.7838 & 0.6598 & 0.7039 & 0.8098 & 0.8416 & \cellcolor[rgb]{ .851,  .851,  .851} 0.5971 & 0.6095 & 0.5371 & 0.9541 & 0.2943 & 0.3099 & 0.9629 & \cellcolor[rgb]{ .851,  .851,  .851} 1.2919 & 3.4100 & 3.6729 & 0.6242 & \cellcolor[rgb]{ .851,  .851,  .851} 0.7307 & 0.3974 & 0.4034 \\
          &       & \textbf{MAE} & 0.6247 & 0.6491 & 0.6436 & 0.7138 & 0.7175 & \cellcolor[rgb]{ .851,  .851,  .851} 0.7631 & \cellcolor[rgb]{ .851,  .851,  .851} 0.6376 & 0.6513 & \cellcolor[rgb]{ .851,  .851,  .851} 0.5393 & 0.8873 & 0.4610 & 0.4848 & \cellcolor[rgb]{ .851,  .851,  .851} 0.7223 & 0.9456 & \cellcolor[rgb]{ .851,  .851,  .851} 1.4190 & 1.5551 & 0.4396 & 0.5119 & 0.4458 & 0.4526 \\
\cmidrule{3-23}          & \multirow{2}[2]{*}{\textbf{iTransformer}} & \textbf{MSE} & \cellcolor[rgb]{ .851,  .851,  .851} 0.7155 & 0.7682 & 0.6477 & 0.7151 & 0.8034 & 0.8350 & 0.6375 & 0.6437 & 0.5477 & 0.9340 & \cellcolor[rgb]{ .851,  .851,  .851} 0.2568 & \cellcolor[rgb]{ .851,  .851,  .851} 0.3004 & 0.9749 & 1.2942 & 3.2786 & 3.5638 & \cellcolor[rgb]{ .851,  .851,  .851} 0.5607 & 0.7564 & 0.3974 & 0.4029 \\
          &       & \textbf{MAE} & \cellcolor[rgb]{ .851,  .851,  .851} 0.6192 & 0.6419 & 0.6364 & 0.7175 & 0.7179 & 0.7634 & 0.6531 & 0.6582 & 0.5472 & 0.8847 & \cellcolor[rgb]{ .851,  .851,  .851} 0.4232 & \cellcolor[rgb]{ .851,  .851,  .851} 0.4451 & 0.7435 & 0.9240 & 1.4324 & 1.5443 & 0.4093 & 0.5766 & \cellcolor[rgb]{ .851,  .851,  .851} 0.4443 & \cellcolor[rgb]{ .851,  .851,  .851} 0.4497 \\
\cmidrule{2-23}          & \multirow{2}[2]{*}{Mean} & MSE   & \cellcolor[rgb]{ .918,  .949,  .973} 0.8622 & \cellcolor[rgb]{ .918,  .949,  .973} 0.9443 & \cellcolor[rgb]{ .918,  .949,  .973} 0.8086 & \cellcolor[rgb]{ .918,  .949,  .973} 0.8540 & \cellcolor[rgb]{ .918,  .949,  .973} 1.2040 & \cellcolor[rgb]{ .918,  .949,  .973} 1.2927 & \cellcolor[rgb]{ .918,  .949,  .973} 0.9769 & \cellcolor[rgb]{ .918,  .949,  .973} 0.9918 & \cellcolor[rgb]{ .918,  .949,  .973} 0.8043 & \cellcolor[rgb]{ .918,  .949,  .973} 1.1084 & \cellcolor[rgb]{ .918,  .949,  .973} 0.3794 & \cellcolor[rgb]{ .918,  .949,  .973} 0.4337 & \cellcolor[rgb]{ .918,  .949,  .973} 1.1148 & \cellcolor[rgb]{ .918,  .949,  .973} 1.3397 & \cellcolor[rgb]{ .918,  .949,  .973} 3.7421 & \cellcolor[rgb]{ .918,  .949,  .973} 3.9974 & \cellcolor[rgb]{ .918,  .949,  .973} 0.9232 & \cellcolor[rgb]{ .918,  .949,  .973} 1.1130 & \cellcolor[rgb]{ .918,  .949,  .973} 0.6476 & \cellcolor[rgb]{ .918,  .949,  .973} 0.6657 \\
          &       & MAE   & \cellcolor[rgb]{ .918,  .949,  .973} 0.7004 & \cellcolor[rgb]{ .918,  .949,  .973} 0.7397 & \cellcolor[rgb]{ .918,  .949,  .973} 0.7478 & \cellcolor[rgb]{ .918,  .949,  .973} 0.8112 & \cellcolor[rgb]{ .918,  .949,  .973} 0.9071 & \cellcolor[rgb]{ .918,  .949,  .973} 0.9744 & \cellcolor[rgb]{ .918,  .949,  .973} 0.8566 & \cellcolor[rgb]{ .918,  .949,  .973} 0.8672 & \cellcolor[rgb]{ .918,  .949,  .973} 0.7182 & \cellcolor[rgb]{ .918,  .949,  .973} 0.9769 & \cellcolor[rgb]{ .918,  .949,  .973} 0.5467 & \cellcolor[rgb]{ .918,  .949,  .973} 0.5940 & \cellcolor[rgb]{ .918,  .949,  .973} 0.8436 & \cellcolor[rgb]{ .918,  .949,  .973} 0.9693 & \cellcolor[rgb]{ .918,  .949,  .973} 1.5589 & \cellcolor[rgb]{ .918,  .949,  .973} 1.6592 & \cellcolor[rgb]{ .918,  .949,  .973} 0.5699 & \cellcolor[rgb]{ .918,  .949,  .973} 0.6739 & \cellcolor[rgb]{ .918,  .949,  .973} 0.6217 & \cellcolor[rgb]{ .918,  .949,  .973} 0.6517 \\
    \midrule
    \midrule
    \multirow{2}[2]{*}{\begin{sideways}\textbf{CNN}\end{sideways}} & \multirow{2}[2]{*}{\textbf{TimesNet}} & \textbf{MSE} & 0.7189 & 0.7515 & 0.6512 & 0.7069 & 0.8168 & 0.8489 & 0.6135 & 0.6263 & 0.5653 & 0.9222 & 0.2857 & 0.3298 & 0.9886 & 1.2988 & 3.3939 & 3.6700 & 0.8146 & 0.9870 & 0.4003 & 0.4063 \\
          &       & \textbf{MAE} & 0.6267 & 0.6391 & 0.6463 & 0.7075 & 0.7490 & 0.7965 & 0.6391 & 0.6529 & 0.5519 & \cellcolor[rgb]{ .851,  .851,  .851} 0.8419 & 0.4720 & 0.5169 & 0.8008 & 0.9657 & 1.4859 & 1.6192 & 0.4859 & 0.5615 & 0.4594 & 0.4663 \\
    \midrule
    \midrule
    \multirow{6}[6]{*}{\begin{sideways}\textbf{Diffusion-based}\end{sideways}} & \multirow{2}[2]{*}{\textbf{TimeGrad}} & \textbf{MSE} & 1.1029 & 1.2754 & 1.0437 & 1.0949 & 1.5287 & 1.6888 & 1.4028 & 1.5319 & 0.9817 & 1.0708 & 0.9628 & 1.2800 & 1.0363 & 1.3211 & 3.8575 & 4.2921 & 1.3107 & 1.6539 & 0.6765 & 0.7049 \\
          &       & \textbf{MAE} & 0.7753 & 0.8463 & 0.8394 & 0.9186 & 1.0550 & 1.1920 & 0.9840 & 1.0252 & 0.7424 & 0.9145 & 0.8345 & 1.0639 & 0.8429 & 1.0130 & 1.5649 & 1.7562 & 0.9082 & 1.1337 & 0.6821 & 0.7291 \\
\cmidrule{3-23}          & \multirow{2}[2]{*}{\textbf{TimeDiff}} & \textbf{MSE} & 0.8023 & 0.8837 & 0.7364 & 0.7684 & 0.9533 & 0.9907 & 0.8598 & 0.8776 & 0.7735 & 1.1308 & 0.6060 & 0.9407 & 1.0412 & 1.3264 & 3.9294 & 4.4300 & 1.2154 & 1.5336 & 0.6127 & 0.6306 \\
          &       & \textbf{MAE} & 0.6883 & 0.7314 & 0.7262 & 0.7710 & 0.8600 & 0.9145 & 0.7869 & 0.8039 & 0.6657 & 0.9647 & 0.6845 & 0.8493 & 0.8699 & 1.0316 & 1.5726 & 1.8444 & 0.8687 & 1.0845 & 0.6478 & 0.6921 \\
\cmidrule{2-23}          & \multirow{2}[2]{*}{Mean} & MSE   & \cellcolor[rgb]{ .918,  .949,  .973} 0.9526 & \cellcolor[rgb]{ .918,  .949,  .973} 1.0796 & \cellcolor[rgb]{ .918,  .949,  .973} 0.8901 & \cellcolor[rgb]{ .918,  .949,  .973} 0.9317 & \cellcolor[rgb]{ .918,  .949,  .973} 1.2410 & \cellcolor[rgb]{ .918,  .949,  .973} 1.3398 & \cellcolor[rgb]{ .918,  .949,  .973} 1.1313 & \cellcolor[rgb]{ .918,  .949,  .973} 1.2048 & \cellcolor[rgb]{ .918,  .949,  .973} 0.8776 & \cellcolor[rgb]{ .918,  .949,  .973} 1.1008 & \cellcolor[rgb]{ .918,  .949,  .973} 0.7844 & \cellcolor[rgb]{ .918,  .949,  .973} 1.1104 & \cellcolor[rgb]{ .918,  .949,  .973} 1.0388 & \cellcolor[rgb]{ .918,  .949,  .973} 1.3238 & \cellcolor[rgb]{ .918,  .949,  .973} 3.8935 & \cellcolor[rgb]{ .918,  .949,  .973} 4.3611 & \cellcolor[rgb]{ .918,  .949,  .973} 1.2631 & \cellcolor[rgb]{ .918,  .949,  .973} 1.5938 & \cellcolor[rgb]{ .918,  .949,  .973} 0.6446 & \cellcolor[rgb]{ .918,  .949,  .973} 0.6678 \\
          &       & MAE   & \cellcolor[rgb]{ .918,  .949,  .973} 0.7318 & \cellcolor[rgb]{ .918,  .949,  .973} 0.7889 & \cellcolor[rgb]{ .918,  .949,  .973} 0.7828 & \cellcolor[rgb]{ .918,  .949,  .973} 0.8448 & \cellcolor[rgb]{ .918,  .949,  .973} 0.9575 & \cellcolor[rgb]{ .918,  .949,  .973} 1.0533 & \cellcolor[rgb]{ .918,  .949,  .973} 0.8855 & \cellcolor[rgb]{ .918,  .949,  .973} 0.9146 & \cellcolor[rgb]{ .918,  .949,  .973} 0.7041 & \cellcolor[rgb]{ .918,  .949,  .973} 0.9396 & \cellcolor[rgb]{ .918,  .949,  .973} 0.7595 & \cellcolor[rgb]{ .918,  .949,  .973} 0.9566 & \cellcolor[rgb]{ .918,  .949,  .973} 0.8564 & \cellcolor[rgb]{ .918,  .949,  .973} 1.0223 & \cellcolor[rgb]{ .918,  .949,  .973} 1.5688 & \cellcolor[rgb]{ .918,  .949,  .973} 1.8003 & \cellcolor[rgb]{ .918,  .949,  .973} 0.8885 & \cellcolor[rgb]{ .918,  .949,  .973} 1.1091 & \cellcolor[rgb]{ .918,  .949,  .973} 0.6650 & \cellcolor[rgb]{ .918,  .949,  .973} 0.7106 \\
    \midrule
    \midrule
    \multirow{12}[12]{*}{\begin{sideways}\textbf{Linear-based}\end{sideways}} & \multirow{2}[2]{*}{\textbf{NLinear}} & \textbf{MSE} & 0.7170 & 0.7819 & 0.6493 & 0.6904 & 0.8212 & \cellcolor[rgb]{ .851,  .851,  .851} 0.8207 & 0.6221 & 0.6350 & \cellcolor[rgb]{ .851,  .851,  .851} 0.5121 & \cellcolor[rgb]{ .851,  .851,  .851} 0.8240 & 0.3015 & 0.3784 & 0.9838 & 1.2984 & \cellcolor[rgb]{ .851,  .851,  .851} 2.9205 & \cellcolor[rgb]{ .851,  .851,  .851} 3.2475 & 0.7761 & 0.9793 & 0.4933 & 0.5161 \\
          &       & \textbf{MAE} & 0.6304 & 0.6638 & 0.6510 & \cellcolor[rgb]{ .851,  .851,  .851} 0.6888 & 0.8231 & 0.8324 & 0.6500 & 0.6640 & 0.5990 & 0.8583 & 0.4715 & 0.5505 & 0.8275 & 0.9896 & 1.4502 & 1.6124 & 0.5315 & 0.6316 & 0.5680 & 0.6281 \\
\cmidrule{3-23}          & \multirow{2}[2]{*}{\textbf{DLinear}} & \textbf{MSE} & 0.7172 & 0.7625 & 0.6426 & 0.7273 & 1.0733 & 1.0830 & 0.6180 & 0.6240 & 0.6498 & 0.9297 & 0.3015 & 0.3783 & 0.9683 & \cellcolor[rgb]{ .851,  .851,  .851} 1.2919 & 3.2040 & 3.5423 & 0.7835 & 0.9885 & 0.4046 & 0.4107 \\
          &       & \textbf{MAE} & 0.6318 & 0.6421 & 0.6373 & 0.7198 & 0.8965 & 0.9317 & 0.6422 & 0.6473 & 0.6714 & 0.9080 & 0.4735 & 0.5518 & 0.8125 & 0.9232 & 1.4962 & 1.5625 & 0.5394 & 0.6395 & 0.5029 & 0.5105 \\
\cmidrule{3-23}          & \multirow{2}[2]{*}{\textbf{TiDE}} & \textbf{MSE} & 0.7813 & 0.8307 & 0.7150 & 0.7878 & 0.8853 & 0.9289 & 0.6405 & 0.6467 & 0.6639 & 1.0043 & 0.3593 & 0.3791 & 0.9818 & 1.2956 & 3.7060 & 3.9686 & 0.9868 & 1.1465 & 0.4103 & 0.4159 \\
          &       & \textbf{MAE} & 0.6584 & 0.6748 & 0.6874 & 0.7692 & 0.7491 & 0.8202 & 0.6609 & 0.6661 & 0.6635 & 0.9415 & 0.5412 & 0.5777 & 0.7980 & 0.9336 & 1.5392 & 1.7068 & 0.6898 & 0.7921 & 0.5029 & 0.5089 \\
\cmidrule{3-23}          & \multirow{2}[2]{*}{\textbf{TimeMixer}} & \textbf{MSE} & 0.7252 & 0.7626 & 0.6577 & 0.7281 & 0.8328 & 0.9239 & 0.6011 & 0.6136 & 0.5406 & 0.9597 & 0.2749 & 0.3449 & \cellcolor[rgb]{ .851,  .851,  .851} 0.9612 & 1.3053 & 3.3377 & 3.6654 & 0.6309 & 0.7645 & \cellcolor[rgb]{ .851,  .851,  .851} 0.3770 & \cellcolor[rgb]{ .851,  .851,  .851} 0.3827 \\
          &       & \textbf{MAE} & 0.6286 & 0.6361 & 0.6487 & 0.7197 & 0.7207 & 0.7636 & 0.6471 & 0.6610 & 0.5506 & 0.8904 & 0.4489 & 0.5230 & 0.7268 & \cellcolor[rgb]{ .851,  .851,  .851} 0.8442 & 1.4293 & \cellcolor[rgb]{ .851,  .851,  .851} 1.5416 & 0.4409 & \cellcolor[rgb]{ .851,  .851,  .851} 0.5095 & 0.4526 & 0.4595 \\
\cmidrule{3-23}          & \multirow{2}[2]{*}{\textbf{TQNet}} & \textbf{MSE} & 0.7174 & \cellcolor[rgb]{ .851,  .851,  .851} 0.7462 & 0.6292 & 0.6946 & \cellcolor[rgb]{ .851,  .851,  .851} 0.7929 & 0.8241 & 0.5985 & \cellcolor[rgb]{ .851,  .851,  .851} 0.5962 & 0.5268 & 0.8836 & 0.2662 & 0.3017 & 0.9821 & 1.3012 & 3.2876 & 3.5875 & 0.5830 & 0.7864 & 0.3779 & 0.3831 \\
          &       & \textbf{MAE} & 0.6229 & \cellcolor[rgb]{ .851,  .851,  .851} 0.6218 & \cellcolor[rgb]{ .851,  .851,  .851} 0.6265 & 0.6951 & \cellcolor[rgb]{ .851,  .851,  .851} 0.7090 & 0.7740 & 0.6426 & \cellcolor[rgb]{ .851,  .851,  .851} 0.6309 & 0.5623 & 0.8607 & 0.4360 & 0.4469 & 0.7980 & 0.9572 & 1.4439 & 1.5668 & \cellcolor[rgb]{ .851,  .851,  .851} 0.4006 & 0.5643 & 0.4666 & 0.4723 \\
\cmidrule{2-23}          & \multirow{2}[2]{*}{Mean} & MSE   & \cellcolor[rgb]{ .918,  .949,  .973} 0.7316 & \cellcolor[rgb]{ .918,  .949,  .973} 0.7768 & \cellcolor[rgb]{ .918,  .949,  .973} 0.6588 & \cellcolor[rgb]{ .918,  .949,  .973} 0.7256 & \cellcolor[rgb]{ .918,  .949,  .973} 0.8811 & \cellcolor[rgb]{ .918,  .949,  .973} 0.9161 & \cellcolor[rgb]{ .918,  .949,  .973} 0.6160 & \cellcolor[rgb]{ .918,  .949,  .973} 0.6231 & \cellcolor[rgb]{ .918,  .949,  .973} 0.5786 & \cellcolor[rgb]{ .918,  .949,  .973} 0.9203 & \cellcolor[rgb]{ .918,  .949,  .973} 0.3007 & \cellcolor[rgb]{ .918,  .949,  .973} 0.3565 & \cellcolor[rgb]{ .918,  .949,  .973} 0.9754 & \cellcolor[rgb]{ .918,  .949,  .973} 1.2985 & \cellcolor[rgb]{ .918,  .949,  .973} 3.2912 & \cellcolor[rgb]{ .918,  .949,  .973} 3.6023 & \cellcolor[rgb]{ .918,  .949,  .973} 0.7521 & \cellcolor[rgb]{ .918,  .949,  .973} 0.9330 & \cellcolor[rgb]{ .918,  .949,  .973} 0.4126 & \cellcolor[rgb]{ .918,  .949,  .973} 0.4217 \\
          &       & MAE   & \cellcolor[rgb]{ .918,  .949,  .973} 0.6344 & \cellcolor[rgb]{ .918,  .949,  .973} 0.6477 & \cellcolor[rgb]{ .918,  .949,  .973} 0.6502 & \cellcolor[rgb]{ .918,  .949,  .973} 0.7185 & \cellcolor[rgb]{ .918,  .949,  .973} 0.7797 & \cellcolor[rgb]{ .918,  .949,  .973} 0.8244 & \cellcolor[rgb]{ .918,  .949,  .973} 0.6486 & \cellcolor[rgb]{ .918,  .949,  .973} 0.6539 & \cellcolor[rgb]{ .918,  .949,  .973} 0.6094 & \cellcolor[rgb]{ .918,  .949,  .973} 0.8918 & \cellcolor[rgb]{ .918,  .949,  .973} 0.4742 & \cellcolor[rgb]{ .918,  .949,  .973} 0.5300 & \cellcolor[rgb]{ .918,  .949,  .973} 0.7926 & \cellcolor[rgb]{ .918,  .949,  .973} 0.9296 & \cellcolor[rgb]{ .918,  .949,  .973} 1.4718 & \cellcolor[rgb]{ .918,  .949,  .973} 1.5980 & \cellcolor[rgb]{ .918,  .949,  .973} 0.5204 & \cellcolor[rgb]{ .918,  .949,  .973} 0.6274 & \cellcolor[rgb]{ .918,  .949,  .973} 0.4986 & \cellcolor[rgb]{ .918,  .949,  .973} 0.5159 \\
    \midrule
    \midrule
    \multirow{2}[2]{*}{\begin{sideways}\textbf{Ours}\end{sideways}} & \multirow{2}[2]{*}{\textbf{T.Magnifier}} & \textbf{MSE} & \cellcolor[rgb]{ .937,  .871,  .804} \textbf{0.6949} & \cellcolor[rgb]{ .937,  .871,  .804} \textbf{0.7029} & \cellcolor[rgb]{ .937,  .871,  .804} \textbf{0.5865} & \cellcolor[rgb]{ .937,  .871,  .804} \textbf{0.5933} & \cellcolor[rgb]{ .937,  .871,  .804} \textbf{0.7859} & \cellcolor[rgb]{ .937,  .871,  .804} \textbf{0.8102} & \cellcolor[rgb]{ .937,  .871,  .804} \textbf{0.4811} & \cellcolor[rgb]{ .937,  .871,  .804} \textbf{0.4531} & \cellcolor[rgb]{ .937,  .871,  .804} \textbf{0.4855} & \cellcolor[rgb]{ .937,  .871,  .804} \textbf{0.7600} & \cellcolor[rgb]{ .937,  .871,  .804} \textbf{0.2185} & \cellcolor[rgb]{ .937,  .871,  .804} \textbf{0.2770} & \cellcolor[rgb]{ .937,  .871,  .804} \textbf{0.9307} & \cellcolor[rgb]{ .937,  .871,  .804} \textbf{1.1526} & \cellcolor[rgb]{ .937,  .871,  .804} \textbf{2.8336} & \cellcolor[rgb]{ .937,  .871,  .804} \textbf{2.0045} & \cellcolor[rgb]{ .937,  .871,  .804} \textbf{0.5262} & \cellcolor[rgb]{ .937,  .871,  .804} \textbf{0.6729} & \cellcolor[rgb]{ .937,  .871,  .804} \textbf{0.2516} & \cellcolor[rgb]{ .937,  .871,  .804} \textbf{0.2527} \\
          &       & \textbf{MAE} & \cellcolor[rgb]{ .937,  .871,  .804} \textbf{0.5222} & \cellcolor[rgb]{ .937,  .871,  .804} \textbf{0.5148} & \cellcolor[rgb]{ .937,  .871,  .804} \textbf{0.5380} & \cellcolor[rgb]{ .937,  .871,  .804} \textbf{0.5305} & \cellcolor[rgb]{ .937,  .871,  .804} \textbf{0.5044} & \cellcolor[rgb]{ .937,  .871,  .804} \textbf{0.6115} & \cellcolor[rgb]{ .937,  .871,  .804} \textbf{0.4706} & \cellcolor[rgb]{ .937,  .871,  .804} \textbf{0.4640} & \cellcolor[rgb]{ .937,  .871,  .804} \textbf{0.5237} & \cellcolor[rgb]{ .937,  .871,  .804} \textbf{0.6241} & \cellcolor[rgb]{ .937,  .871,  .804} \textbf{0.3251} & \cellcolor[rgb]{ .937,  .871,  .804} \textbf{0.3727} & \cellcolor[rgb]{ .937,  .871,  .804} \textbf{0.5529} & \cellcolor[rgb]{ .937,  .871,  .804} \textbf{0.6862} & \cellcolor[rgb]{ .937,  .871,  .804} \textbf{1.3950} & \cellcolor[rgb]{ .937,  .871,  .804} \textbf{1.2740} & \cellcolor[rgb]{ .937,  .871,  .804} \textbf{0.3425} & \cellcolor[rgb]{ .937,  .871,  .804} \textbf{0.3746} & \cellcolor[rgb]{ .937,  .871,  .804} \textbf{0.3023} & \cellcolor[rgb]{ .937,  .871,  .804} \textbf{0.3043} \\
    \bottomrule
    \end{tabular}%
    }
    \par\vspace{0.6mm}\parbox{0.98\linewidth}{\footnotesize\raggedright\textit{Note:} T.Magnifier is short for TimeMagnifier. Mean rows average the models of each category and are not ranked.}\par
  \label{tab: main}%
  \vspace{-0.4cm}
\end{table*}%

\subsection{Datasets and Evaluation Metrics}
To assess TimeMagnifier's generalization across diverse scenarios, we use ten benchmark datasets spanning environment, energy, economics, healthcare, and transportation \cite{shao2023exploring,wang2024deep,qiu2024tfb}. Several are themselves Web-published rollups, as the hourly Electricity series aggregates 15-minute UCI readings and Traffic uses hourly PeMS aggregates \cite{lai2018modeling}. The descriptions and statistical properties of these datasets are presented in \textbf{Appendix~\ref{app:datasets}}. To ensure consistency with prior studies, we employ Mean Squared Error (MSE) and Mean Absolute Error (MAE) as the primary evaluation metrics, complemented by CRPS, interval coverage, and band-wise spectral diagnostics in \textbf{Appendix~\ref{app:spectral}}.

\subsection{Baselines Description}
To evaluate TimeMagnifier in super-resolution forecasting, we compare it with 15 state-of-the-art (SOTA) and widely used long-term time series forecasting (LTSF) baselines. These baselines cover diverse architectures, ensuring a fair and representative comparison. We group them into four categories: 1) \textbf{Transformer-based models:} Transformer \cite{vaswani2017attention}, Informer \cite{zhou2021informer}, Autoformer \cite{wu2021autoformer}, FEDformer \cite{zhou2022fedformer}, Crossformer \cite{zhang2023crossformer}, PatchTST \cite{nietime}, iTransformer \cite{liu2024itransformer}. 2) \textbf{CNN-based model:} TimesNet \cite{wu2022timesnet}. 3) \textbf{Diffusion-based models:} TimeGrad \cite{rasul2021autoregressive}, TimeDiff \cite{shen2023non}. 4) \textbf{Linear-based models:} NLinear \cite{zeng2023transformers}, DLinear \cite{zeng2023transformers}, TiDE \cite{daslong}, TimeMixer \cite{wang2024timemixer}, TQNet \cite{lin2025temporal}. Implementation details are provided in \textbf{Appendix~\ref{app:baselines}}.

\subsection{Experimental Setup}
All experiments are conducted on a unified computing platform equipped with a 2.70 GHz Intel® Xeon® CPU Max 9462 processor and two NVIDIA A100 80GB SXM GPUs. We train with Adam at learning rate $5\times10^{-4}$ and batch size 64 for at most 100 epochs, with early stopping at patience 3 on the validation set and learning-rate halving after any epoch without validation gain, and all methods, baselines included, share a 20-trial search. Splits are chronological, 12/4/4 months for ETT and 7:1:2 otherwise, and are made on the native grid before downsampling, so no future fine value leaks. Main results report the mean of three runs; for every cell, TimeMagnifier and the runner-up are additionally rerun with five seeds, and these five-seed means, which differ slightly from the three-run values, underlie the significance tests in \textbf{Appendix~\ref{app:stats}}.

\subsection{Main Results}
We set $\Gamma = 2$ and $\Gamma = 4$ with $\gamma = 1$, a look-back window of 168 coarse steps, and a forecast horizon of 336, or a look-back of 24 and horizon of 36 for Illness.
\textbf{Table~\ref{tab: main}} reports the super-resolution forecasting results. TimeMagnifier is best in every cell, and \textbf{Appendix~\ref{app:stats}} shows that over five seeds its gain over the runner-up passes a Diebold-Mariano test at $p<0.01$ in 19 of 20 MSE settings, all but the short and noisy Illness at $\Gamma=2$. At $\Gamma = 2$ it reduces MSE over the strongest baseline by 0.9\% to 33.3\%, with the largest margins of 33.3\% on Weather and 19.4\% on ETTm1, while the MAE reductions exceed 25\% on ETTh2, ETTm1, and Weather, reflecting smaller typical errors. The margin also widens as inputs grow coarser on nine of ten datasets, since the relative MSE gain over the runner-up rises with $\Gamma$ from 5.9\% to 12.9\% on ETTh1, from 19.4\% to 24.0\% on ETTm1, and from 3.2\% to 10.8\% on ExRate, so a dedicated coarse-to-fine design matters more as the gap grows.
No baseline is consistently competitive, suggesting that no standard LTSF architecture is consistently suited to MTS-SRF. At $\Gamma = 2$, iTransformer and TimeMixer come closest, together runner-up on five datasets, while FEDformer is strong on ETTh1, whose periodic structure suits its frequency-domain decomposition. NLinear takes second place on ETTm2 and Illness at both coefficients, whereas Transformer and Informer perform worst, as they also do in standard same-resolution forecasting. At $\Gamma = 4$ coarser inputs make forecasting harder, as 147 of 150 baseline MSE entries degrade and the other three improve by under 1\%; even iTransformer and TQNet lose about 18\% on average, while PatchTST and TimesNet occasionally reach second best, for instance on Traffic and on the MAE of ETTm2, suggesting that patching and multi-scale designs partially mitigate the sparsity of coarse observations. TimeMagnifier improves in MSE at $\Gamma = 4$ on Illness and ETTm1, possibly in part because the look-back, counted in coarse steps, then covers at least one full annual or weekly cycle, 96 rather than 48 weeks and 7 rather than 3.5 days.
\begin{table}[t]
  \setlength\tabcolsep{4pt}
  \centering
  \caption{MSE of baseline adapters at $\Gamma=4$; P1 to P4 use the same splits and 20-trial budget as TimeMagnifier.}
  \vspace{-0.3cm}
  \resizebox{0.96\linewidth}{!}{
  \begin{tabular}{l|cc|cc}
  \toprule
  \multirow{2}{*}{Adapter} & \multicolumn{2}{c|}{ETTh1} & \multicolumn{2}{c}{Weather} \\
   & TQNet & iTrans. & TQNet & iTrans. \\
  \midrule
  P0 linear head (Table~\ref{tab: main}) & 0.6946 & 0.7151 & 0.3831 & 0.4029 \\
  P1 fine-grid head & 0.7038 & 0.7165 & 0.3880 & 0.4088 \\
  P2 forecast, then spline & 0.7330 & 0.7353 & 0.4041 & 0.4173 \\
  P3 spline, then forecast & 0.7190 & 0.7308 & 0.3916 & 0.4166 \\
  P4 sub-pixel upsampler & 0.6935 & 0.7152 & 0.3889 & 0.4053 \\
  \midrule
  mTAN & \multicolumn{2}{c|}{0.7132} & \multicolumn{2}{c}{0.3941} \\
  TimeMagnifier & \multicolumn{2}{c|}{\textbf{0.5933}} & \multicolumn{2}{c}{\textbf{0.2527}} \\
  \bottomrule
  \end{tabular}}
  \label{tab:protocol}
  \vspace{-0.4cm}
\end{table}

\subsection{Strengthened Baseline Protocol}\label{subsec:protocol}
LTSF baselines are not built for MTS-SRF, so \textbf{Table~\ref{tab: main}} equips each with a learned linear head that maps its $h_G$ coarse steps to $h$ fine steps and contains linear and nearest-neighbor interpolation as special cases. To test whether this head handicaps them, we give TQNet and iTransformer four alternatives at $\Gamma=4$ with the same splits and 20-trial budget: P1, a head that outputs the fine grid directly; P2, forecasting followed by cubic-spline interpolation; P3, spline upsampling of the history followed by forecasting; and P4, a parameter-matched sub-pixel upsampler from image super-resolution \cite{shi2016real}. We also evaluate mTAN \cite{shukla2021multi}, a continuous-time model queried at fine-grid timestamps. In \textbf{Table~\ref{tab:protocol}}, the 16 new configurations range from 0.2\% better to 5.5\% worse than the linear head, and TimeMagnifier is 14.4\% and 34.0\% below the best adapters, P4 on ETTh1 and the original head P0 on Weather.

\subsection{Study on Horizon and Look-back Window}

We fix $\Gamma = 2$, $\gamma = 1$, and a look-back of 168. Forecast horizons are set to 96, 192, 336, and 729, where 729 is indivisible by $\Gamma=2$ and thus exercises the block retention of Eq.~\eqref{eq:compatible_horizon} with $h=730$ and one unscored step. For the short Illness set, the look-back is 24 and the four rows use horizons 12, 24, 36, and 60. \textbf{Table~\ref{tab:horizon}} reports results across all ten datasets. MSE and MAE rise monotonically with the horizon on every dataset, most steeply on the short Illness, whose MSE grows about 13 times from the shortest to the longest horizon, followed by ExRate at 3.3 times and ETTm2 at 2.8 times, while the other datasets grow by at most about two times.

\begin{figure}[t]
  \centering
  \includegraphics[width=1\linewidth]{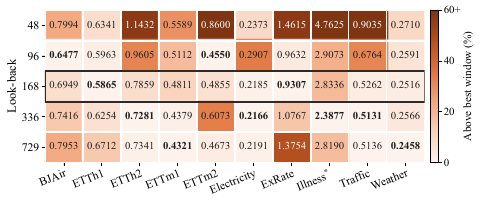}
  \vspace{-0.6cm}
 \caption{Look-back study at $\Gamma=2$ and horizon 336. MSE is shaded by its gap to each dataset's best window (bold). $^{*}$Illness uses horizon 36 and windows 12, 18, 24, 36 and 48.}
  \label{fig:window}
  \vspace{-0.3cm}
\end{figure}

\noindent\textbf{Look-back window.} With the horizon fixed, \textbf{Figure~\ref{fig:window}} varies the look-back over 48, 96, 168, 336, and 729, or 12 to 48 for Illness. Error is lowest at an intermediate length on eight datasets, as short windows lack context and long ones add noise, while ETTm1 and Weather do best at 729. The shared look-back of 168 in main results is best for TimeMagnifier on only two datasets, so its gains do not rest on a tuned window.

\begin{table}[!t]
\setlength\tabcolsep{1pt}
  \centering
  \caption{Results across different forecast horizons.}
   \vspace{-0.3cm}
  \resizebox{1\linewidth}{!}{
    \begin{tabular}{ccc|cc|cc|cc|cc}
    \toprule
    \multirow{2}[4]{*}{Horizon} & \multicolumn{2}{c}{BJAir} & \multicolumn{2}{c}{ETTh1} & \multicolumn{2}{c}{ETTh2} & \multicolumn{2}{c}{ETTm1} & \multicolumn{2}{c}{ETTm2} \\
\cmidrule{2-11}        & MSE & MAE & MSE & MAE & MSE & MAE & MSE & MAE & MSE & MAE \\
    \midrule
    96  & 0.5144  & 0.4372  & 0.4341  & 0.4504  & 0.4661  & 0.4709  & 0.3519  & 0.3754  & 0.2202  & 0.3123  \\
    192 & 0.5822  & 0.4707  & 0.4914  & 0.4850  & 0.5363  & 0.4906  & 0.3919  & 0.4197  & 0.3607  & 0.4100  \\
    336 & 0.6949  & 0.5222  & 0.5865  & 0.5380  & 0.7859  & 0.5044  & 0.4811  & 0.4706  & 0.4855  & 0.5237  \\
    729 & 0.8882  & 0.6437  & 0.7496  & 0.6633  & 0.8469  & 0.5143  & 0.5560  & 0.5343  & 0.6132  & 0.6288  \\
    \midrule
    \multirow{2}[4]{*}{Horizon} & \multicolumn{2}{c|}{Electricity} & \multicolumn{2}{c|}{ExRate} & \multicolumn{2}{c|}{Illness} & \multicolumn{2}{c|}{Traffic} & \multicolumn{2}{c}{Weather} \\
\cmidrule{2-11}        & MSE & MAE & MSE & MAE & MSE & MAE & MSE & MAE & MSE & MAE \\
    \midrule
    96  & 0.1693  & 0.2768  & 0.3851  & 0.3267  & 0.2367  & 0.4267  & 0.4450  & 0.2997  & 0.1575  & 0.2216  \\
    192 & 0.1872  & 0.2936  & 0.7429  & 0.4249  & 1.5579  & 1.0661  & 0.5067  & 0.3365  & 0.2028  & 0.2669  \\
    336 & 0.2185  & 0.3251  & 0.9307  & 0.5529  & 2.8336  & 1.3950  & 0.5262  & 0.3425  & 0.2516  & 0.3023  \\
    729 & 0.2549  & 0.3470  & 1.2753  & 0.7324  & 3.0369  & 1.5966  & 0.6297  & 0.3703  & 0.3268  & 0.3569  \\
    \bottomrule
    \end{tabular}%
    }
  \label{tab:horizon}%
  \vspace{-0.3cm}
\end{table}%

\begin{table}[t]
  \setlength\tabcolsep{3pt}
  \centering
  \caption{Comparison of baseline linear adaptation and De-noise Only \& DTW plug-in adaptation for the MTS-SRF task.}
    \vspace{-0.3cm}
  \resizebox{0.95\linewidth}{!}{
  \begin{tabular}{cccc|cc|cc}
  \toprule
  \multirow{2}[4]{*}{Models} & \multirow{2}[4]{*}{Metrics} & \multicolumn{2}{c}{ETTh1} & \multicolumn{2}{c}{Electricity} & \multicolumn{2}{c}{Traffic} \\
\cmidrule{3-8}  &  & $\Gamma=2$ & $\Gamma=4$ & $\Gamma=2$ & $\Gamma=4$ & $\Gamma=2$ & $\Gamma=4$ \\
  \midrule
  \multirow{2}{*}{DLinear} & MSE & 0.6426 & 0.7273 & 0.3015 & 0.3783 & 0.7835 & 0.9885 \\
   & MAE & 0.6373 & 0.7198 & 0.4735 & 0.5518 & 0.5394 & 0.6395 \\
  \midrule
  \multirow{2}{*}{PatchTST} & MSE & 0.6598 & 0.7039 & 0.2943 & 0.3099 & 0.6242 & 0.7307 \\
   & MAE & 0.6436 & 0.7138 & 0.4610 & 0.4848 & 0.4396 & 0.5119 \\
  \midrule
  DLinear & MSE & 0.5988 & 0.6619 & 0.2849 & 0.3589 & 0.7393 & 0.9054 \\
  w/ D\&D & MAE & 0.5779 & 0.6547 & 0.4474 & 0.5245 & 0.4927 & 0.5824 \\
  \midrule
  PatchTST & MSE & 0.6144 & 0.6503 & 0.2761 & 0.2901 & 0.5885 & 0.6848 \\
  w/ D\&D & MAE & 0.5882 & 0.6598 & 0.4283 & 0.4544 & 0.4097 & 0.4797 \\
  \bottomrule
  \end{tabular}}
  \label{tab:plugin}
    \vspace{-0.5cm}
\end{table}

\subsection{Ablation Study}
\begin{figure*}[t]
\vspace{-0.5cm}
    \centering
    \begin{subfigure}[t]{0.22\textwidth}
        \centering
        \includegraphics[width=\linewidth]{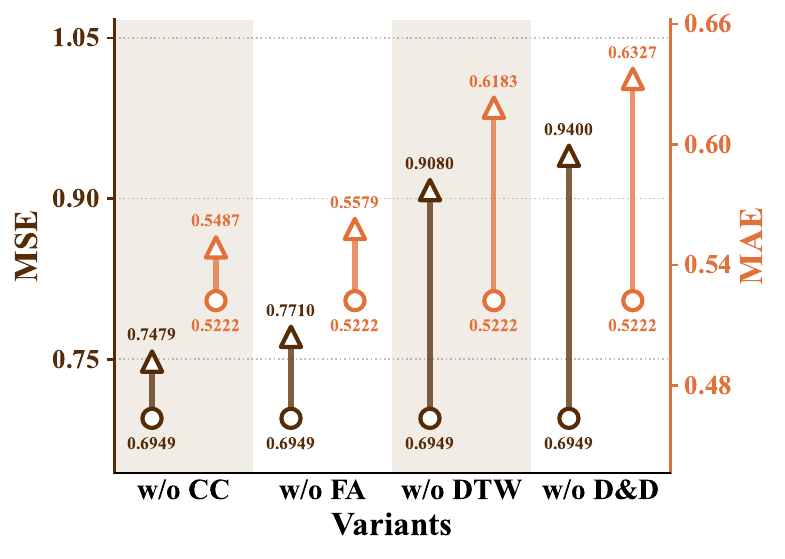}
        \vspace{-0.6cm}
        \caption{BJAir Dataset}
    \end{subfigure}
    \hfill
    \begin{subfigure}[t]{0.22\textwidth}
        \centering
        \includegraphics[width=\linewidth]{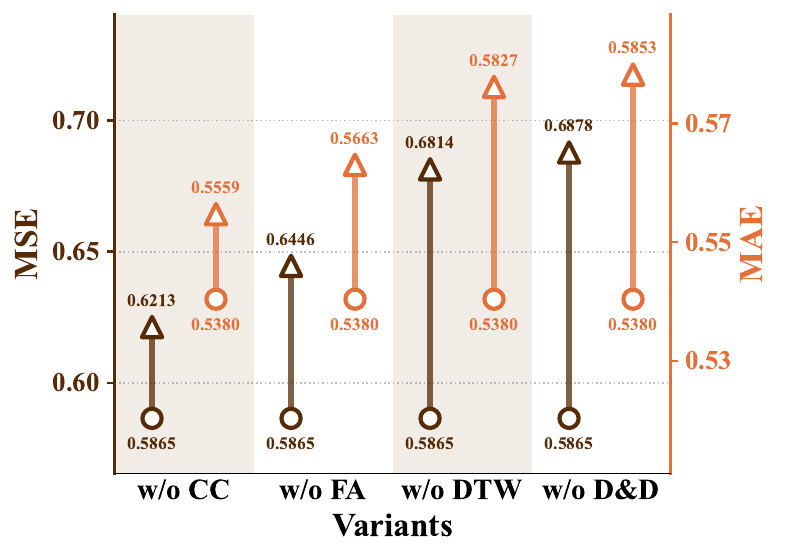}
        \vspace{-0.6cm}
        \caption{ETTh1 Dataset}
    \end{subfigure}
    \hfill
    \begin{subfigure}[t]{0.22\textwidth}
        \centering
        \includegraphics[width=\linewidth]{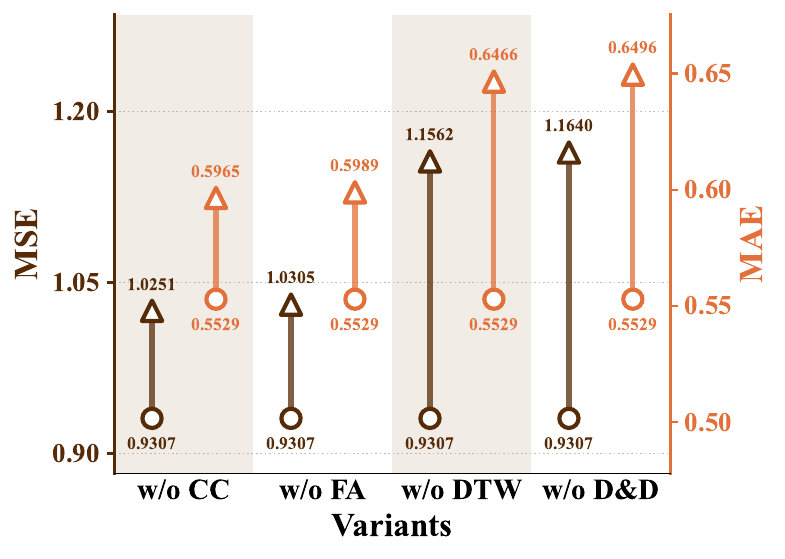}
        \vspace{-0.6cm}
        \caption{ExRate Dataset}
    \end{subfigure}
    \hfill
    \begin{subfigure}[t]{0.22\textwidth}
        \centering
        \includegraphics[width=\linewidth]{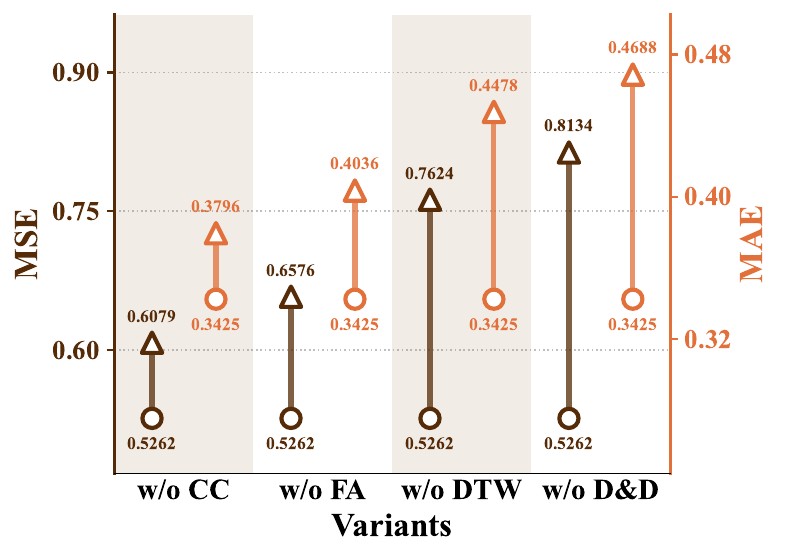}
        \vspace{-0.6cm}
        \caption{Traffic Dataset}
    \end{subfigure}
    \hfill
\begin{minipage}[c]{0.08\textwidth}
        \centering
        \vspace{-2.6cm} %
        \includegraphics[width=\linewidth]{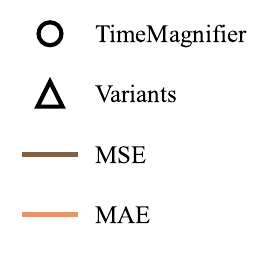}
    \end{minipage}
        \vspace{-0.4cm}
    \caption{Ablation study results. Longer lines indicate greater degradation in predictive accuracy.}
    \label{fig:ablation}
    \vspace{-0.5cm}
\end{figure*}
Under the main settings at $\Gamma=2$ and $\gamma=1$, we remove one component at a time and retune the rest under the full model's 20-trial budget, since inherited hyperparameters would confound removal with poor tuning. The variants are \textbf{w/o~CC}, without cross-variable interaction; \textbf{w/o~FA}, a shared linear coarse forecaster without gates or periodic prior; \textbf{w/o~DTW}, without alignment; and \textbf{w/o~D\&D}, replacing De-noise Only and DTW by a linear output layer. On all ten datasets and both metrics, the ordering is TimeMagnifier $<$ w/o~CC $<$ w/o~FA $<$ w/o~DTW $<$ w/o~D\&D, with MSE increases of 5.5\% to 54.6\%, largest on Traffic, Electricity, and ETTm2; \textbf{Figure~\ref{fig:ablation}} shows four of the ten datasets. Removing DTW alone already accounts for 77\% to 97\% of the degradation caused by removing both stages, so level-wise refinement drifts without alignment. Fourier Anchor with a bare linear head trails DLinear at 0.6878 vs. 0.6426, as expected of the low-capacity prior analyzed in \textbf{Appendix~\ref{app:projection}}, whose value emerges once refinement is added, as \textbf{Section~\ref{subsec:plugin}} shows.

\subsection{Study on Resolution Coefficients}
\begin{figure}[t]
  \centering
  \vspace{0.2cm}
  \begin{subfigure}[t]{0.49\linewidth}\centering
    \includegraphics[width=\linewidth]{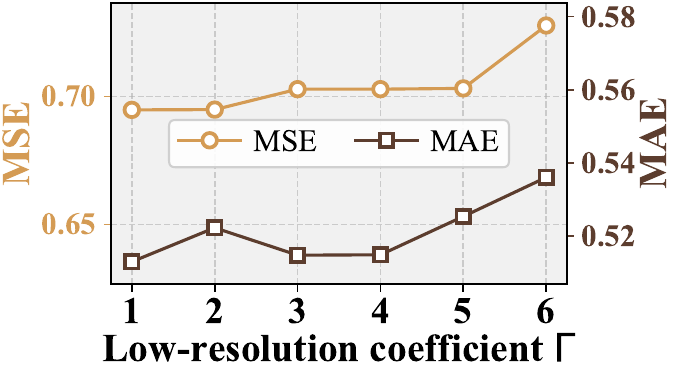}
    \vspace{-0.55cm}\caption{$\Gamma$ on BJAir}
  \end{subfigure}\hfill
  \begin{subfigure}[t]{0.49\linewidth}\centering
    \includegraphics[width=\linewidth]{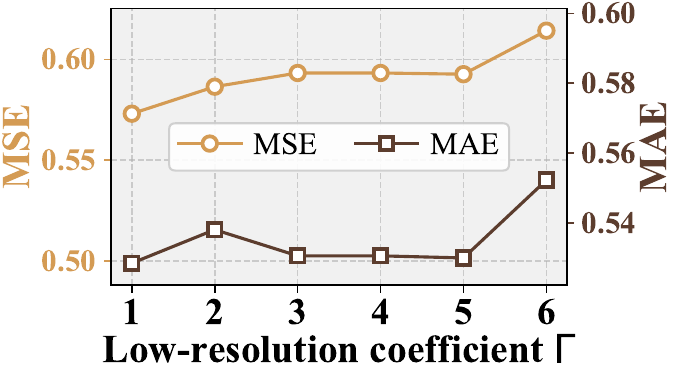}
    \vspace{-0.55cm}\caption{$\Gamma$ on ETTh1}
  \end{subfigure}\\[-0.1cm]
  \begin{subfigure}[t]{0.49\linewidth}\centering
    \includegraphics[width=\linewidth]{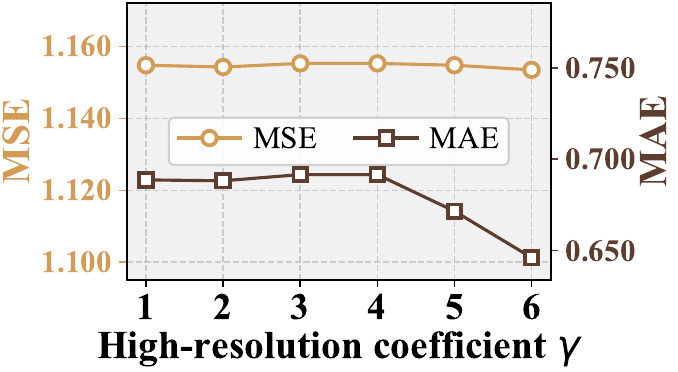}
    \vspace{-0.55cm}\caption{$\gamma$ on ExRate}
  \end{subfigure}\hfill
  \begin{subfigure}[t]{0.49\linewidth}\centering
    \includegraphics[width=\linewidth]{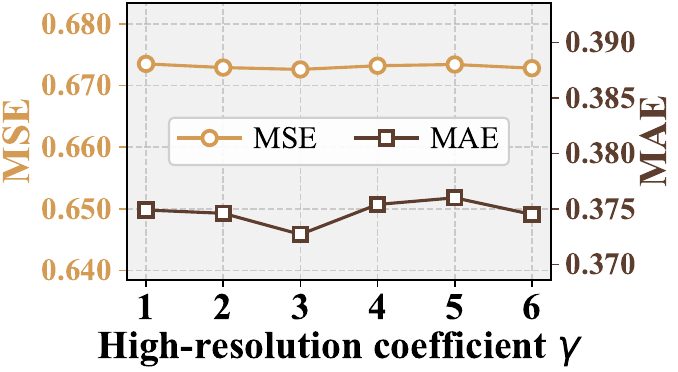}
    \vspace{-0.55cm}\caption{$\gamma$ on Traffic}
  \end{subfigure}
  \caption{Resolution study: $\Gamma$ with $\gamma=1$ (a, b) and $\gamma$ with $\Gamma=6$ (c, d); $\Gamma=1$ and $\gamma=6$ are same-resolution controls.}
  \label{fig:resolution}
  \vspace{-0.3cm}
\end{figure}

\textbf{Figure~\ref{fig:resolution}} varies $\Gamma$ from 1 to 6 with $\gamma=1$, and $\gamma$ from 1 to 6 with $\Gamma=6$ over a fixed physical horizon, where $\Gamma=1$ and $\gamma=6$ are same-resolution controls. Error grows gently with $\Gamma$ on both BJAir and ETTh1, by about 5\% in MSE from $\Gamma=2$ to $\Gamma=6$, so accuracy decays gracefully as archives coarsen. Across $\gamma$, MSE varies by under 0.3\% on ExRate and Traffic, even at the control $\gamma=6$, so at $\Gamma=6$ a finer grid adds little error on these sets.

\subsection{De-noise Only \& DTW Plug-in Analysis}\label{subsec:plugin}

We examine whether the De-noise Only \& DTW components (D\&D) can serve as a plug-in that gives existing LTSF baselines MTS-SRF capability. Under the main settings, we append D\&D to DLinear and PatchTST, keeping their tuned backbones and hyperparameters; since these expose no Fourier coefficients, the plug-in estimates them from each input window by the ridge regression of Eq.~\eqref{eq:conditional_Fourier Anchor}, so it carries the full periodic prior. D\&D reduces MSE by 5.1\% to 9.0\% in all twelve cases of \textbf{Table~\ref{tab:plugin}}. Because the plug-in and the full model share D\&D and its prior, their gap isolates the coarse base, and the full model leads the best plug-in in all six settings by 1.7\% to 20.9\% in MSE. Base and refinement also interact: without D\&D, Fourier Anchor trails DLinear on Traffic at 0.8134 vs. 0.7835, but with D\&D it leads at 0.5262 vs. 0.7393, a crossover that additive effects cannot produce.

\subsection{Further Analyses}\label{subsec:further}
\textbf{Appendices~\ref{app:spectral} to \ref{app:dtwalt}} extend the evaluation beyond point accuracy. On ETTm1, TimeMagnifier stays ahead of TQNet when coarse inputs are mean-aggregated, 30\% missing, or jittered, and unlike TQNet it adds no excess power above the coarse Nyquist frequency. On ETTh1, it reaches a CRPS of 0.48 with 91\% coverage, against 0.54 to 0.58 for the diffusion baselines. Its learned alignment weights also outperform Euclidean, cosine, and Soft-DTW alignment, with Soft-DTW needing 2.22 times our runtime.

\section{Related Work}
MTS-SRF jointly assumes coarse-only inputs at inference, a finer future grid, and fine data only as training supervision; the lines of work below each relax at least one of these assumptions.

\noindent\textbf{Mixed-Frequency and Cross-Resolution Modeling.}
Mixed-data sampling (MIDAS) regressions \cite{ghysels2007midas} predict low-frequency targets from high-frequency regressors, and reverse MIDAS \cite{foroni2018using} predicts high-frequency variables but still conditions on their high-frequency lags, whereas MTS-SRF observes no fine value at inference. Temporal disaggregation \cite{chow1971best,denton1971adjustment} distributes observed past totals over finer periods and can extrapolate using fine-grained indicators. Temporal hierarchies \cite{athanasopoulos2017forecasting} reconcile forecasts across aggregation levels but never forecast below the observed resolution. Imputation and continuous-time models such as mTAN \cite{shukla2021multi} interpolate between observed points, and we evaluate mTAN under our protocol in Section~\ref{subsec:protocol}. Climate downscaling \cite{vandal2017deepsd} and audio super-resolution \cite{kuleshov2017audio} share the coarse-to-fine goal but super-resolve a given span rather than extrapolate a future one.

\noindent\textbf{Multi-Scale Time Series Forecasting.}
Multi-scale models exploit dependencies across scales through graphs \cite{chen2023multi}, iterative refinement \cite{shabani2022scaleformer}, adaptive pathways \cite{chen2024pathformer}, warping of irregular series \cite{zhang2023warpformer}, or MLP mixing \cite{zhong2023multi,hu2024adaptive}. They rescale only internally and observe at the native resolution, so the fine data that MTS-SRF lacks at inference remain available to them.

\noindent\textbf{Diffusion for Time Series Forecasting.}
Diffusion models corrupt data with noise and learn to reverse the process. TimeGrad \cite{rasul2021autoregressive} pioneered diffusion for forecasting, TimeDiff \cite{shen2023non} stabilized denoising, and ScoreGrad \cite{yan2021scoregrad} introduced score matching, all at a fixed resolution with costly multi-step sampling. Closer to our setting, Cold Diffusion \cite{bansal2023cold} replaces noise with deterministic degradations such as downsampling, and mr-Diff \cite{shen2024multi} denoises from coarse to fine trends. Both keep a fixed-length grid and learn a denoising network, whereas De-noise Only lengthens the grid at every level, supervises each level with resampled fine targets, injects per-window periodic priors, and adds only two scalar parameters.

\vspace{-0.1cm}
\section{Conclusion and Future Work}

In this study, we introduce Multivariate Time Series Super-Resolution Forecasting (MTS-SRF), a new task that predicts high-resolution sequences from low-resolution inputs, addressing fine-grained forecasting from coarse archives in Web telemetry, energy, and environmental monitoring. To overcome data sparsity, fine-grained ambiguity, and cross-resolution misalignment, we propose TimeMagnifier, which couples a Fourier Anchor stage, a progressive De-noise Only stage, and a DTW alignment stage. Extensive experiments confirm that TimeMagnifier outperforms SOTA LTSF models across diverse datasets, and its De-noise Only and DTW components further transfer as a portable plug-in that improves existing forecasters. MTS-SRF also provides a foundation for large-scale pretraining and multi-resolution time series integration. Promising directions include self-supervised pretraining that obtains cross-resolution supervision without paired data, and adaptive strategies that adjust granularity to data characteristics.

\bibliographystyle{ACM-Reference-Format}
\bibliography{sample-base}

\appendix

\begin{table}[h]
  \setlength\tabcolsep{2pt} 
  \centering
  \vspace{-0.4cm}
  \caption{Dataset statistics.}
  \vspace{-0.3cm}
  \resizebox{0.85\linewidth}{!}{
    \begin{tabular}{ccccc}
    \toprule
    Datasets & Timestamps & Channels & Freq.($\Delta \tau_{\text{min}}$) & Domain \\
    \midrule
    BJAir & 36,000 & 9   & 1 hour & Environment \\
    ETTh1 & 14,400 & 7   & 1 hour & Energy \\
    ETTh2 & 14,400 & 7   & 1 hour & Energy \\
    ETTm1 & 57,600 & 7   & 15 min & Energy \\
    ETTm2 & 57,600 & 7   & 15 min & Energy \\
    Electricity & 26,304 & 321 & 1 hour & Energy \\
    ExRate & 7,588 & 8   & 1 day & Economics \\
    Illness & 966 & 7   & 1 week & Healthcare \\
    Traffic & 17,544 & 862 & 1 hour & Transportation \\
    Weather & 52,696 & 21  & 10 min & Environment \\
    \bottomrule
    \end{tabular}%
    }
  \label{tab:Dataset}%
  \vspace{-0.6cm}
\end{table}%

\section{Dataset Description \label{app:datasets}}
 
We utilize ten benchmark multivariate time series datasets across environment, energy, economics, healthcare, and transportation. \textbf{Table~\ref{tab:Dataset}} lists their statistics.
 
\section{Baseline Setup \label{app:baselines}}
All baseline models are obtained from publicly available benchmark implementations for time series forecasting\footnote{https://github.com/decisionintelligence/TFB}\textsuperscript{,}\footnote{https://github.com/GestaltCogTeam/BasicTS} or from their official repositories. To ensure fairness and reproducibility, we reproduce and evaluate all models in a unified experimental environment.
 
None of the baselines has an upsampling path, so we reshape each baseline's $h_G$ coarse outputs into $h$ fine steps with the learned linear layer of \textbf{Algorithm~\ref{alg:baseline_srp}}, which preserves every backbone and contains linear and nearest-neighbor interpolation as special cases. \textbf{Section~\ref{subsec:protocol}} compares this head with four alternative adapters and mTAN under the same 20-trial budget. For the plug-in of \textbf{Table~\ref{tab:plugin}}, the Fourier coefficients needed by the periodic correction are estimated from each input window by the ridge regression of Eq.~\eqref{eq:conditional_Fourier Anchor}, since DLinear and PatchTST do not produce them.
 
\noindent\textbf{Hyperparameter settings.} We set $K = 3$. For each dataset, the three dominant physical periods $\Pi_k$ are taken from the periodogram of the coarse training split, shared by all variables, and kept fixed for training and testing; at a given $\Gamma$ each spans $\Pi_k/(\Gamma\Delta\tau_{\text{min}}) > 2$ coarse steps, below the coarse Nyquist frequency, e.g., 12 steps at $\Gamma=2$ and 6 at $\Gamma=4$ for the daily period of hourly data. The smoothing coefficient \( \alpha \) and the periodic attenuation \( \beta \) are initialized to 0.5 and learned during training, and the alignment coefficient \( \eta = \sigma(\xi) \) is initialized with \( \xi = 0 \). We initialize $s_1 = s_2 = 0$ and $s_3 = \log 2$, corresponding to $\lambda_1 = \lambda_2 = 1.0$ and $\lambda_3 = 0.5$, and learn them during training. The DTW alignment uses the scaled-diagonal band of Eq.~\eqref{eq:dtw_band} with $B = \lceil r \rceil$. The ridge coefficient of Eq.~\eqref{eq:conditional_Fourier Anchor} is fixed to $\lambda_{\mathrm{ridge}} = 10^{-3}$ for all datasets. The cross-variable residual strength $\kappa$ is initialized to 0 and learned during training. Neighborhoods use $|\mathcal{N}_i| = \min(16, N-1)$ and are computed once from the training split.
 
\begin{algorithm}[t]
\caption{Baseline Model Adaptation}
\label{alg:baseline_srp}
\KwIn{ 
    Low-resolution coefficient $\Gamma$, 
    High-resolution coefficient \( \gamma \), 
    Look-back window length $T$, 
    Forecast horizon $h_{\mathrm{req}}$,
    Coarse-grained time series data $X \in \mathbb{R}^{T \times N}$
}
\KwOut{
    Fine-grained prediction $\hat{Y} \in \mathbb{R}^{h_{\mathrm{req}} \times N}$
}
 
\BlankLine
\textbf{Step 1: Baseline Model Prediction}\\
$\hat{Y}_{\mathrm{coarse}} \gets \text{BaselineModel}(X)$ \tcp*[l]{Baseline model predicts $h_G$ coarse steps}
 
\BlankLine
\textbf{Step 2: Linear Upsampling for Super-Resolution}\\
$\hat{Y} \gets \text{LinearLayer}(\hat{Y}_{\mathrm{coarse}})$ \tcp*[l]{Linear transformation to adjust output length to $h$}
 
\BlankLine
\Return the first $h_{\mathrm{req}}$ steps of $\hat{Y}$
\end{algorithm}

\section{Statistical Reliability \label{app:stats}}
For every cell of \textbf{Table~\ref{tab: main}}, we rerun TimeMagnifier and the runner-up with five seeds on the fixed chronological splits and apply a two-sided Diebold-Mariano test with HAC variance to the seed-averaged per-window errors. \textbf{Table~\ref{tab:seeds}} lists the four $\Gamma=2$ cells with the smallest relative gaps in \textbf{Table~\ref{tab: main}}. The first three are significant ($p<0.01$) with TimeMagnifier's standard deviation at most 2.0\% of its mean (at most 1.7\% for the runner-ups). Illness is not ($p=0.255$), as its 966 weekly points leave few test windows.
\begin{table}[h]
  \setlength\tabcolsep{3pt}
  \centering
  \caption{Smallest-gap cells over five seeds ($\Gamma=2$, MSE).}
  \vspace{-0.3cm}
  \resizebox{1\linewidth}{!}{
  \begin{tabular}{l|c|lc|c}
  \toprule
  Dataset & TimeMagnifier & \multicolumn{2}{c|}{Runner-up in Table~\ref{tab: main}} & $p$ \\
  \midrule
  ETTh2 & 0.781 $\pm$ 0.008 & TQNet & 0.799 $\pm$ 0.009 & 0.0015 \\
  BJAir & 0.690 $\pm$ 0.006 & iTransformer & 0.717 $\pm$ 0.011 & 0.0021 \\
  ExRate & 0.950 $\pm$ 0.019 & TimeMixer & 0.980 $\pm$ 0.017 & 0.0027 \\
  Illness & 2.806 $\pm$ 0.149 & NLinear & 2.922 $\pm$ 0.154 & 0.255 \\
  \bottomrule
  \end{tabular}}
  \label{tab:seeds}
  \vspace{-0.4cm}
\end{table}
 
\section{Spectral, Probabilistic, and Anchor Checks \label{app:spectral}}
\noindent\textbf{Band-wise recovery.} The periodic prior injects only modes below the coarse Nyquist frequency $f_N$, and each level smooths once, so content below $f_N$ is recoverable in principle up to aliasing from higher bands, while content above $f_N$ is not identifiable from coarse samples alone. \textbf{Table~\ref{tab:bands}} tests this on ETTm1 ($\Gamma=4$) with the ratio $\rho$ of predicted to true power and the coherence $C$. Above $f_N$, TimeMagnifier attenuates power ($\rho=0.81$) but has low coherence ($C=0.16$), while TQNet injects excess power with lower coherence.
\begin{table}[h]
  \setlength\tabcolsep{4pt}
  \centering
  \vspace{-0.3cm}
  \caption{Band-wise power ratio $\rho$ and coherence $C$ on ETTm1 ($\Gamma=4$).}
  \vspace{-0.3cm}
  \begin{tabular}{l|cc|cc}
  \toprule
  Band & TM $\rho$ & TM $C$ & TQNet $\rho$ & TQNet $C$ \\
  \midrule
  below $0.5 f_N$ & 0.95 & 0.79 & 1.05 & 0.54 \\
  $0.5 f_N$ to $f_N$ & 0.91 & 0.69 & 1.31 & 0.45 \\
  above $f_N$ & 0.81 & 0.16 & 1.12 & 0.08 \\
  \bottomrule
  \end{tabular}
  \label{tab:bands}
  \vspace{-0.3cm}
\end{table}
 
\noindent\textbf{Probabilistic evaluation.} TimeMagnifier is a point forecaster, so an empirical residual construction forms its predictive distribution by adding the empirical validation residuals, per variable and horizon step, to the point forecast; CRPS is computed from this distribution and the 90\% interval from its 5\% and 95\% quantiles. On ETTh1 ($\Gamma=2$), TimeMagnifier attains CRPS 0.48 and 91\% empirical coverage, versus 0.58 and 88\% for TimeGrad and 0.54 and 86\% for TimeDiff, which under-cover.
 
\noindent\textbf{Anchor bias.} On ETTh1 ($\Gamma=2$) the learned alignment step is $\eta=0.23$, so alignment corrects only partially. Replacing $\bar{Y}$ by an oracle anchor built from the true future coarse values lowers MSE only from 0.5865 to 0.5541, and on the 25\% of windows with the largest anchor error alignment still lowers MSE by 0.03.
 
\section{Robustness to Acquisition Artifacts \label{app:robust}}
We train on clean point-sampled data and test on ETTm1 ($\Gamma=4$) with perturbed coarse inputs. TimeMagnifier versus the best baseline, TQNet, obtains 0.4923 versus 0.5857 under mean aggregation, 0.5296 versus 0.6474 with 30\% of coarse points missing and forward filled, and 0.5142 versus 0.6034 with timestamp jitter within the coarse interval (MSE; clean inputs give 0.4531 versus 0.5962).
 
\section{Alignment Alternatives \label{app:dtwalt}}
\textbf{Table~\ref{tab:dtwalt}} replaces our alignment by simpler alternatives at $\Gamma=4$. Banded DTW, Euclidean keeps our band, path search, and correction but weights all matched pairs equally, and shared path, cosine finds one banded path per sample and level for all channels from cosine distances of cross-channel vectors. All alternatives improve over no alignment but trail ours, and Soft-DTW, the closest on Traffic, needs 2.22 times the runtime. Since the Euclidean variant and ours share the path search and the correction of Section~\ref{subsec:dtw_correct}, their gap comes from $\nu$ alone. A uniform $\mathcal{L}_{\text{DTW}}$ pulls every fine step toward the stepwise anchor, including in-between steps where the anchor is a poor target, and so works against $\mathcal{L}_{\text{refine}}$; $\nu$ lets training relax this pull where the two disagree and keep it where the anchor is reliable, consistent with Remark~\ref{prop:dtwbound}.
\begin{table}[h]
  \setlength\tabcolsep{4pt}
  \centering
  \vspace{-0.3cm}
  \caption{Alignment alternatives ($\Gamma=4$; MSE, time ratio).}
  \vspace{-0.3cm}
    \resizebox{0.82\linewidth}{!}{
  \begin{tabular}{l|cc|c}
  \toprule
  Alignment & ETTh1 & Traffic & Time (Traffic) \\
  \midrule
  None & 0.7157 & 0.7953 & 0.81 \\
  Banded DTW, Euclidean & 0.6965 & 0.7257 & 1.03 \\
  Shared path, cosine & 0.6701 & 0.7604 & 0.95 \\
  Soft-DTW, same band & 0.6761 & 0.7139 & 2.22 \\
  Ours & \textbf{0.5933} & \textbf{0.6729} & 1.00 \\
  \bottomrule
  \end{tabular}}
  \label{tab:dtwalt}
  \vspace{-0.5cm}
\end{table}
 
\section{Complexity Analysis \label{app:complexity}}
 
We describe the computational cost of one forward pass of TimeMagnifier in terms of the number of variables $N$, the coarse-grid length $T$, the internal horizon $h$, the refinement ratio $r = \Gamma/\gamma$ between coarse and fine resolutions, and the number of Fourier modes $K$.
 
\textbf{Fourier Anchor.} The ridge projection matrix depends only on $\lambda_{\mathrm{ridge}}$ and the frequencies, so it is precomputed. Fitting $C(X)$ and extrapolating it to $h_G$ future coarse steps cost $O(NK(T + h_G))$, while the sparse shared gates on the future coarse grid cost $O(N |\mathcal{N}_i| h_G)$ for a fixed hidden width, adding only $\theta_g$ and $\kappa$.
 
\textbf{De-noise Only.} This stage uses local one-dimensional stencils and the reused Fourier coefficients. At level $j$ the smoothing update and the periodic correction act on the expanded grid of length $L^{(j+1)}$, and both are element-wise (up to nearest-neighbor accesses and $K$ sinusoids), so the cost is $O(N L^{(j+1)} K)$. The expanded lengths follow a doubling schedule $2h_G, 4h_G, \dots$ capped at $h$, so the total refined length satisfies $\sum_{j=0}^{J-1} L^{(j+1)} = O(h)$ and is dominated by the final grid. The overall cost of De-noise Only is therefore $O(NKh)$, independent of $r$, which enters only through the number of levels $J = O(\log r)$. The stage reuses the Fourier coefficients from Fourier Anchor and adds only two scalar parameters ($\alpha$ and $\beta$), so its parameter count does not grow with $T$, $h$, or $r$.
 
\textbf{DTW Alignment.} The band in Eq.~\eqref{eq:dtw_band} assigns at most $2B+1$ refined candidates to each of the $h_G$ coarse indices, so the dynamic programming table has $O(h_G B)$ admissible cells per variable. With $B = \lceil r \rceil$ and $h = r h_G$, the per-level cost is $O(N h_G B) = O(Nh)$, and since alignment runs at all $J = \max(1, \lceil \log_2 r \rceil)$ levels plus one final pass, the total is $O(NhJ)$.
 
\textbf{Overall complexity.} Putting the three stages together, the cost of one forward pass of TimeMagnifier is
\begin{equation}
\text{cost}_{\text{TM}} = O\!\left( NK(T + h_G) + N |\mathcal{N}_i| h_G \right) + O(NKh) + O(NhJ).
\end{equation}
Here $J=\max(1,\lceil \log_2 r\rceil)$ is the number of resolution levels: refinement is linear in the final horizon $h$, while banded alignment is repeated over the $J$ levels. TimeMagnifier thus avoids the quadratic sequence-length cost of self-attention and the multi-step sampling of diffusion models; its 0.18M parameters do not grow with $h$, and narrowing the band $B$ trades accuracy for speed.

\begin{figure}[t]
  \centering
  \includegraphics[width=0.9\linewidth]{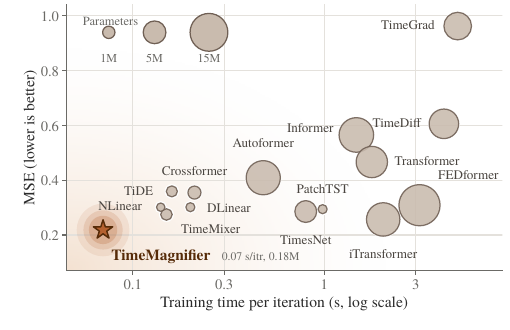}
  \vspace{-0.6cm}
  \caption{Training time per iteration and MSE on Electricity at $\Gamma=2$, with bubble area proportional to trainable parameters; all models use the same GPU and batch size.}
  \label{fig:efficiency}
  \vspace{-0.6cm}
\end{figure}

\noindent\textbf{Measured cost.} \textbf{Figure~\ref{fig:efficiency}} compares training cost, model size, and accuracy on Electricity at $\Gamma=2$. TimeMagnifier needs 0.07 s per iteration, no more than NLinear and DLinear, with 0.18M parameters, behind only these two in size, while both diffusion models need over 4 s per iteration.
 
\section{Recoverability and Challenge Analysis \label{app:feasibility}}
 
\noindent\textbf{Signal Model and Downsampling.}
Let $X = \{ X^{1, \Delta\tau_G}, X^{2, \Delta\tau_G}, \dots,\\ X^{T, \Delta\tau_G} \}$ denote the coarse-grained historical observations with sampling granularity $\Delta\tau_G = \Gamma \Delta\tau_{\text{min}}$. In MTS-SRF, we assume the low-resolution input is obtained from the underlying high-resolution sequence through a deterministic downsampling operator $\mathcal{D}_\Gamma(\cdot)$:
\begin{equation}
X^{t, \Delta\tau_G} = \mathcal{D}_\Gamma \big( X^{t, \Delta\tau_{\text{min}}} \big).
\end{equation}
 
\noindent\textbf{Sources of Recoverable Structure.}
MTS-SRF is not exactly identifiable, since coarse sampling removes or aliases components above the coarse Nyquist limit, so TimeMagnifier does not claim to recover arbitrary missing detail. Instead it exploits three complementary sources that constrain the plausible fine-grained futures: dependencies among variables, low-dimensional trend and periodic structure, and conditional patterns learned from paired training data. The \textbf{w/o CC} and \textbf{w/o FA} ablations probe the first two directly.
 
\noindent\textbf{Challenge 1: Information Loss.}
Downsampling lowers the Nyquist limit to $f^{LR}_{\text{Nyquist}} = \tfrac{1}{2\Gamma\Delta\tau_{\text{min}}}$, so higher-frequency components are lost or aliased into lower frequencies. Fourier Anchor models the low-frequency trend and periodic structure supported by the coarse observations and uses cross-variable correlations to constrain plausible fine-grained forecasts.
 
\noindent\textbf{Challenge 2: Fine-Grained Ambiguity.}
Many fine-grained futures are compatible with the same coarse history, so MTS-SRF estimates a plausible future rather than uniquely inverting $\mathcal{D}_\Gamma$. De-noise Only refines through $J = \max(1,\lceil\log_2(\Gamma/\gamma)\rceil)$ levels, smoothing each intermediate grid and adding the coarse-scale periodic prior. This provides architectural regularization rather than a formal conditioning guarantee.
 
\noindent\textbf{Challenge 3: Temporal Misalignment.}
Coarse-to-fine correspondence is one-to-many and can induce phase shifts. DTW searches constrained warping paths under an unweighted cost and applies level-wise corrections, with learned weights only in the loss, to stay consistent with the coarse forecast.

\begin{tmremark}[Single-level alignment bound]\label{prop:dtwbound}
At level $j$, let $Z^{(j)} = \hat{Y}^{(j,3)}$ be the pre-alignment sequence on $\mathcal{G}^{(j+1)}$, let $t_x^{(j)}(n)$ be the coarse index aligned to $n$ along the DTW path, and let $\hat{Y}^{(j+1)}(n) = Z^{(j)}(n) + \eta \big( \bar{Y}^{t_x^{(j)}(n)} - Z^{(j)}(n) \big)$ with $\eta \in (0,1)$. Then for the target $Y^{(j+1)}$ on $\mathcal{G}^{(j+1)}$,
\begin{equation}
\big\| \hat{Y}^{(j+1)} - Y^{(j+1)} \big\|_2 \le (1-\eta) \big\| Z^{(j)} - Y^{(j+1)} \big\|_2 + \eta \big\| \bar{Y} \circ t_x^{(j)} - Y^{(j+1)} \big\|_2,
\end{equation}
where $(\bar{Y} \circ t_x^{(j)})(n) = \bar{Y}^{t_x^{(j)}(n)}$ is the warped coarse prior.
\end{tmremark}
 
\noindent\textit{Proof.} By definition, $\hat{Y}^{(j+1)} - Y^{(j+1)} = (1-\eta) \big( Z^{(j)} - Y^{(j+1)} \big) + \eta \big( \bar{Y} \circ t_x^{(j)} - Y^{(j+1)} \big)$, and applying the triangle inequality yields the bound. \hfill$\square$
 
The bound shows that alignment interpolates between the refinement error and the discrepancy of the warped coarse prior: it can never amplify the error beyond the worse of the two terms, and it strictly reduces the error whenever the coarse prior read along the warping path is closer to the target than the refined sequence, which is the refinement-drift regime the module targets.
 
\section{Structured Representation of Fourier Anchor \label{app:projection}}
For each input $X$, Eq.~\eqref{eq:conditional_Fourier Anchor} gives $\widetilde{Y}(X)=\Phi_F C(X)$, so its columns lie in the trend-Fourier subspace $\operatorname{col}(\Phi_F)$ of dimension at most $D=2+2K$. Ridge regularization with $\lambda_{\mathrm{ridge}}>0$ keeps the solve for $C(X)$ unique, and, as $P_{\mathrm{FA}}$ is a fixed linear map, the base forecast is Lipschitz in $X$. The data-adaptive cross-variable gates can move the final mixed forecast $\bar{Y}$ outside this subspace.
 
\section{Analysis of De-noise Only \label{app:denoise}}
 
Under the replicated boundary of Eq.~\eqref{eq:smoothing_boundary}, $\mathbf{L}$ is symmetric positive semidefinite with eigenvalues $0 = \zeta_1 \le \dots \le \zeta_{L^{(j+1)}} \le 2$, so the eigenvalues of $S_\alpha = I - \alpha \mathbf{L}$ are $\mu_k = 1 - \alpha \zeta_k$.
 
\begin{tmremark}[Single-pass stability]\label{prop:converge}
For any $z \in \mathbb{R}^{L^{(j+1)}}$ and any $\alpha \in (0,1)$,
\begin{equation}
\left\| S_\alpha z \right\|_2 \le \left\| z \right\|_2 ,
\end{equation}
and $S_\alpha$ strictly damps every non-constant eigenmode of $\mathbf{L}$.
\end{tmremark}
 
\noindent\textit{Proof.} Write $\mathbf{L} = V \operatorname{diag}(\zeta_k) V^\top$, so $S_\alpha = V \operatorname{diag}(\mu_k) V^\top$. Since $0 \le \zeta_k \le 2$ and $\alpha \in (0,1)$, we have $|\mu_k| \le 1$ for every $k$, with strict inequality whenever $\zeta_k > 0$. Hence $\|S_\alpha\|_2 \le 1$. \hfill$\square$
 
This characterizes the operation actually implemented, namely exactly one pass per level: smoothing never amplifies the sequence and strictly attenuates non-constant modes.
 
\section{Applicability and Extensions \label{app:limitations}}
 
\noindent\textbf{Supervision is a training-time requirement, not a deployment constraint.} TimeMagnifier uses paired coarse-fine sequences only during training, and at inference it consumes coarse history alone, matching settings where fine-grained collection is limited by cost, storage, or sensor budget. Such pairs are readily obtained, since any high-resolution segment retained before rollup yields supervision by downsampling. The $\Gamma=4$ ETTm results are real point-sampled archives, but they do not cover sensors that aggregate. Under the mean aggregation, 30\% missing coarse points, and timestamp jitter of \textbf{Appendix~\ref{app:robust}}, TimeMagnifier still leads the best baseline on ETTm1 by 15.9\%, 18.2\%, and 14.8\%, yet its error rises more (8.7\%, 16.9\%, 13.5\%), so acquisition mismatch remains a limitation. The modular loss in Eq.~\eqref{eq:total} also makes the framework compatible with self-supervised extensions, because the refinement term is the only component that consults fine-grained targets, and it can be replaced by a consistency objective that re-downsamples the refined output and matches it against the future coarse targets.
 
\noindent\textbf{Graceful degradation under widening resolution gaps.} The Nyquist analysis in \textbf{Appendix~\ref{app:feasibility}} shows that content above the coarse Nyquist limit is not identifiable from coarse samples alone, so the recoverable bandwidth generally decreases as $\Gamma$ grows, and what matters is how gracefully it decays. Our $\Gamma$ study in \textbf{Figure~\ref{fig:resolution}} shows that TimeMagnifier degrades smoothly up to $\Gamma = 6$, and \textbf{Table~\ref{tab: main}} shows that its advantage over the strongest baseline widens as inputs become coarser on nine of ten datasets. This is consistent with the design, because the trend-Fourier subspace and the cross-variable gates model the predictable structure that survives aggressive downsampling. 
For still larger gaps, the architecture accommodates external evidence without modification, since domain priors or physical constraints can be injected through the periodic prior of Eq.~\eqref{eq:periodic_corr} alongside the learned Fourier coefficients.

\end{document}